%% file: main.tex
\documentclass{article} 
\usepackage{iclr2026_conference,times}

\input{math_commands.tex}

\usepackage{hyperref}
\usepackage{url}
\usepackage{graphicx}
\usepackage{booktabs}
\usepackage{amsmath}   
\usepackage{amssymb}   
\usepackage{amsfonts}  
\usepackage{amsthm}  
\usepackage{amsthm} 

\makeatletter
\@ifundefined{lemma}{%
  \newtheorem{lemma}{Lemma}
}{}
\makeatother

\title{Jacobian Aligned Random Forests}

\author{Sarwesh Rauniyar \\
  Department of Applied Mathematics and Statistics \\
  Johns Hopkins University \\
  Baltimore, Maryland \\ 
  \texttt{srauniy1@jh.edu}
}

\iclrfinalcopy 
\begin{document}

\pagestyle{plain}

\maketitle

\begin{abstract}
Axis-aligned decision trees are fast and stable but struggle on datasets with rotated or interaction-dependent decision boundaries, where informative splits require linear combinations of features rather than single-feature thresholds. Oblique forests address this with per-node hyperplane splits, but at added computational cost and implementation complexity. We propose a simple alternative: JARF, Jacobian-Aligned Random Forests. Concretely, we first fit an axis-aligned forest to estimate class probabilities or regression outputs, compute finite-difference gradients of these predictions with respect to each feature, aggregate them into an expected Jacobian outer product that generalizes the expected gradient outer product (EGOP), and use it as a single global linear preconditioner for all inputs. This supervised preconditioner applies a single global rotation of the feature space, then hands the transformed data back to a standard axis-aligned forest, preserving off-the-shelf training pipelines while capturing oblique boundaries and feature interactions that would otherwise require many axis-aligned splits to approximate. The same construction applies to any model that provides gradients, though we focus on random forests and gradient-boosted trees in this work. On tabular classification and regression benchmarks, this preconditioning consistently improves axis-aligned forests and often matches or surpasses oblique baselines while improving training time. Our experimental results and theoretical analysis together indicate that supervised preconditioning can recover much of the accuracy of oblique forests while retaining the simplicity and robustness of axis-aligned trees.
\end{abstract}

\section{Introduction}
On tabular data, tree-based ensemble methods are widely used and often outperform deep networks on structured datasets \citep{Breiman2001,Grinsztajn2022}. Methods like Random Forests and gradient boosting are popular for their strong performance with minimal tuning, robustness to irrelevant features, and inherent handling of mixed data types. However, these models are fundamentally built on \emph{axis-aligned} decision trees, where each split considers only a single feature. This design makes training fast, but it fails when the boundary depends on a rotated axis or a mix of features. In such cases, an axis-aligned tree must simulate an oblique split through a series of orthogonal cuts, resulting in deeper trees and fragmented decision regions. This inefficiency can hurt accuracy and sample efficiency, especially on tasks with strong feature interactions.

Researchers have long recognized this limitation and explored \emph{oblique} decision trees that split on linear combinations of features rather than single features. Oblique Random Forest variants have shown improved accuracy over standard forests by capturing feature interactions at each node \citep{Menze2011,Katuwal2020}. Unfortunately, these benefits come with significant drawbacks. Learning the optimal linear combination at each node is a more complex optimization problem, often requiring iterative techniques or convex solvers that augment training cost \citep{Murthy1994,Menze2011,Katuwal2020}. Oblique splits also tend to introduce many more parameters and can be prone to overfitting without careful regularization. As a result, oblique forests are often slower and less practical to use than standard axis-aligned ones. 

In this paper, we propose a different way to capture oblique structure: a global, supervised feature transformation that preconditions decision forests. We term our method JARF, short for \emph{Jacobian Aligned Random Forest}. JARF learns a mapping of the input features by leveraging information from a surrogate model’s predicted class probabilities or regression outputs. In particular, we estimate the \emph{expected Jacobian outer product} (EJOP) of the class probability function, which is a covariance-like matrix that measures how sensitive the predictions are to changes in each input direction \citep{Trivedi2014}. For regression, the same construction reduces to the expected gradient outer product (EGOP) of a scalar regression function. By rotating and scaling the original feature space along these EJOP/EGOP directions, JARF creates a new feature space where the most label-predictive variations are axis-aligned. A standard Random Forest or gradient-boosted tree model trained on this transformed space can then simulate oblique decision boundaries using simple axis-aligned splits. The transformation is one-pass and model-agnostic: it requires only a reasonably accurate, smooth surrogate that supplies gradients, after which any tree ensemble can be trained on the transformed inputs without changing its internal training pipeline.

Conceptually, this places JARF at the intersection of supervised projection methods and oblique forests. Like recent average-gradient-outer-product constructions, JARF builds a global positive semidefinite matrix from prediction gradients and uses it to define a new feature geometry, but here the matrix is specialized to the conditional mean or class-probability map and is used as an explicit input-space preconditioner for tree ensembles rather than as a step in a neural representation. On the theoretical side, we analyze how EJOP-based preconditioning changes CART itself. For squared-loss CART, we show that an axis-aligned split in the preconditioned space corresponds to an oblique split in the original space, with impurity gain controlled by the EJOP along the split direction. We further prove that a finite-difference EJOP estimator built from a surrogate forest recovers the same geometric object as classical EJOP under mild smoothness assumptions, and we establish a dimension-adapted risk bound in a ridge-function setting, where JARF behaves like an \(r\)-dimensional nonparametric regressor even when the ambient dimension is much larger. These results go beyond prior EJOP work by tying gradient-based geometry directly to tree impurity gains and by providing guarantees tailored to tree ensembles.

Empirically, we demonstrate that applying JARF closes much of the accuracy gap between axis-aligned and oblique forests. Our evaluation spans 15 diverse classification datasets and 5 regression benchmarks, ranging from medium-scale UCI and OpenML tasks to larger and higher-dimensional problems such as HIGGS and JANNIS. Across this suite, JARF achieves better accuracy than significantly more complex oblique-tree ensembles with substantially lower computational overhead, and also outperforms lighter, data-agnostic oblique variants such as random-rotation and random-projection forests. We further compare against simple global linear projections, including PCA and LDA, and find that EJOP/EGOP-based preconditioning consistently yields stronger forests than these unsupervised or purely discriminative transforms. Finally, we show that JARF can be paired with both Random Forests and gradient-boosted trees, and that its performance is robust to the choice of surrogate family used to estimate EJOP/EGOP. Taken together, these results highlight supervised, gradient-based preconditioning as a simple and general way to endow axis-aligned tree ensembles with much of the representational power of oblique forests while retaining their speed, robustness, and ease of use.

\section{Related Work}
\label{sec:related}

\subsection{Supervised projection for dimension reduction.}
Early work in statistics introduced \emph{supervised} linear projections to reduce dimensionality while preserving predictive information. Sliced Inverse Regression (SIR; \citealp{Li1991}) and Sliced Average Variance Estimation (SAVE; \citealp{Cook2000SAVE}) seek a low-dimensional subspace of features that most influences the response. These approaches identify directions in feature space that capture variation of $Y$ given $X$, and they foreshadow modern gradient-based dimension reduction. Conceptually, they motivate using label information to precondition the inputs before fitting a model, which is a perspective we adopt. For classification, including multiclass, SIR and SAVE apply directly by slicing on class labels \citep{Li1991,Cook2000SAVE}. Closely related, Fisher’s linear discriminant analysis and its multiclass extension (Rao) learn at most one fewer projection than the number of classes, because only that many independent directions are needed to separate the classes \citep{Fisher1936,Rao1948}. In contrast, JARF may use more than $C - 1$ directions when this improves the fit of tree ensembles, since it serves as a global preconditioner rather than a strict dimension reduction step.

\subsection{Gradient-based global sensitivity (EJOP).}
More recent methods leverage derivatives of a predictive function with respect to inputs to find informative projections. In regression, the expected gradient outer product (EGOP) is
\[
\mathbb{E}_{X}\!\big[\nabla f(X)\nabla f(X)^\top\big]
\]
and recovers an effective dimension-reduction subspace \citep{Trivedi2014}. For multiclass settings, the \emph{expected Jacobian outer product} (EJOP) is
\[
\mathbb{E}_{X}\!\big[Jf(X)\,Jf(X)^\top\big],
\]
where $f$ returns class probabilities; the leading eigenvectors emphasize directions along which predictions change the most \citep{Trivedi2020}. EGOP appears as a scalar-output special case of EJOP, so a single EJOP framework covers both regression and classification. Prior work has mainly used these matrices for tasks such as metric learning, representation learning, and sensitivity analysis \citep{Perronnin2010,Sobol2009,Trivedi2014,Trivedi2020}, with theory expressed in terms of risk bounds for the underlying regression or classification function.

Our approach, JARF, follows this gradient-based paradigm but applies it in a different way. We compute an EJOP/EGOP matrix from a surrogate model and then use it as a global linear preconditioner \emph{before} training a tree ensemble. This leads to two differences from previous EJOP work. First, we specialize the construction to tree ensembles and analyze how EJOP-based preconditioning alters axis-aligned CART splitting: for squared loss, we show that a split in the transformed space corresponds to an oblique split in the original space, with impurity gain controlled by EJOP along the split direction. Second, we study a finite-difference EJOP estimator that uses a forest surrogate and prove that it recovers the same geometric object as classical EJOP under mild smoothness conditions, together with a dimension-adapted risk bound in a ridge-function setting where EJOP identifies a low-dimensional predictive subspace even when the ambient dimension is large.

\subsection{Oblique decision forests.}
Decision trees that split on linear combinations of features were shown early on to yield compact, accurate models when boundaries are tilted relative to the axes \citep{Breiman2001}. \emph{OC1} performs hill-climbing at each node to optimize a hyperplane split, trading extra per-node computation for improved fit \citep{Murthy1994}.  \emph{Rotation Forest} applies unsupervised PCA-based rotations to random feature subsets independently per tree, so subsequent axis-aligned splits behave like oblique splits in the original space \citep{Rodriguez2006}. \emph{Canonical Correlation Forests (CCF)} compute supervised projections at each node via canonical correlation with the outputs, aligning splits with local predictive structure \citep{Rainforth2015}. Another line samples random linear combinations for candidate splits. Breiman noted this idea in early forest variants \citep{Breiman2001}, and \emph{Sparse Projection Oblique Random Forests (SPORF)} constrain projections to be very sparse, which improves interaction capture while mitigating overfitting \citep{Tomita2020}. While effective, these methods either increase \emph{per-node} optimization (OC1, CCF) or rely on \emph{unsupervised or random} projections (Rotation Forest, SPORF) that do not necessarily align with predictive directions. This often means more trees or extra constraints are needed to match the performance of well tuned axis-aligned forests.

\subsection{Comparison and positioning of JARF.}
Unlike node-wise oblique methods, JARF provides a \emph{one-pass}, \emph{global}, and \emph{supervised} preconditioning that leaves the tree learner unchanged. By constructing a single EGOP/EJOP-based transform shared across all trees, JARF supplies a coherent feature representation informed by all training labels, with negligible overhead during tree construction. This global projection amplifies directions along which $p(y\mid x)$ varies and attenuates irrelevant ones so that standard axis-aligned splits can approximate oblique boundaries. In this sense, JARF occupies a middle ground between simple global linear projections such as PCA or LDA and fully oblique forests: it uses label information to shape a global geometry, but keeps the underlying learner as an off-the-shelf axis-aligned forest or gradient-boosted ensemble.

From another perspective, JARF resembles a one-step random feature model. We first learn a smooth surrogate and extract an EJOP/EGOP matrix from its gradients, then freeze this geometry and train a separate tree ensemble on the preconditioned inputs. This two-stage view separates the representation learning from the final predictor, which allows JARF to work with any surrogate that provides gradients (for example neural networks, kernel methods, or tree ensembles with finite differences) and with any downstream tree learner. To the best of our knowledge, prior EJOP and EGOP work has not analyzed how such gradient-based preconditioning changes CART impurity gains, and prior oblique forests have not used EJOP-style sensitivity matrices to define a single supervised global preconditioner for all trees.

\section{Methods}
\label{sec:method}

\subsection{Problem setup and notation}
We consider multiclass classification with inputs $x \in \mathbb{R}^d$ and labels $y \in \{1,\dots,C\}$. Let $f:\mathbb{R}^d \!\to\! \Delta^{C-1}$ denote a probabilistic classifier whose $c$-th component $f_c(x)$ estimates $p(y=c\mid x)$. Standard Random Forests (RF; \citealp{Breiman2001}) and gradient-boosted trees build axis-aligned decision trees on $X\!=\![x_1,\dots,x_n]^\top$, which can require deep trees when informative directions are linear combinations of features. Our goal is to learn a single, global, supervised linear map $H \in \mathbb{R}^{d\times d}$ such that training an ordinary tree ensemble on the transformed data $XH$ captures those predictive combinations with shallow, axis-aligned splits. The same construction applies to regression with real-valued labels $y \in \mathbb{R}$ by replacing class probabilities with a scalar regression function.

\subsection{Probability–gradient preconditioning}
The central object in JARF is an EJOP-style matrix that summarizes how predictions change with small perturbations of $x$. Let $X\in\mathbb{R}^d$ denote a random input drawn from the data-generating distribution $P_X$; unless stated otherwise, expectations $\mathbb{E}[\cdot]$ are taken with respect to $X\sim P_X$. Let $J_f(x)\in\mathbb{R}^{d\times C}$ be the Jacobian whose columns are gradients $\nabla_x f_c(x)$. The \emph{expected Jacobian outer product (EJOP)} is
\begin{equation}
\label{eq:ejop}
H_0 \;\;=\;\; \mathbb{E}_{X}\!\big[J_f(X)J_f(X)^\top\big]
\;\;=\;\; \sum_{c=1}^{C}\,\mathbb{E}_{X}\!\big[\nabla_x f_c(X)\,\nabla_x f_c(X)^\top\big],
\end{equation}
a positive semidefinite matrix whose leading eigenvectors span directions along which $p(y\!\mid\!x)$ varies most \citep{Trivedi2014,Trivedi2020}. For regression tasks with scalar output $y \in \mathbb{R}$, we instead take $f:\mathbb{R}^d \to \mathbb{R}$ to be the regression function and recover the expected gradient outer product (EGOP)
\[
H_0 \;=\; \mathbb{E}_{X}\!\big[\nabla f(X)\nabla f(X)^\top\big],
\]
which appears as the $C=1$ special case of \eqref{eq:ejop}. In both cases, $H_0$ defines a global, label informed geometry on the input space.

JARF uses this geometry as a preconditioner for tree ensembles. In practice, we replace $\mathbb{E}_{X}$ by an empirical average over a subsample of the training inputs to estimate $H_0$, then build a linear map $H$ from this estimate and apply it to all features before training the forest. The tree learner itself remains unchanged: it still searches over axis-aligned splits, but in a feature space where directions that strongly affect $p(y\mid x)$ have been amplified and rotated toward coordinate axes.

\paragraph{Connection to supervised dimension reduction.}
Equation \eqref{eq:ejop} is the gradient or Jacobian analogue of supervised projection methods such as SIR and SAVE \citep{Li1991,Cook2000SAVE}: instead of relying on first or second moments of $X\mid Y$, JARF aggregates the sensitivity of $p(y\!\mid\!x)$ to $x$, producing a label informed geometry. Unlike classical dimension-reduction methods that are often restricted to at most $C-1$ directions in a $C$-class problem, we treat $H_0$ as a full preconditioner and allow the downstream tree ensemble to decide how many directions to exploit.

\subsection{Estimating $H_0$ via finite differences}
The estimator below follows the EJOP construction of \citet{Trivedi2020}, but uses a tree based surrogate rather than a kernel estimator. We construct an empirical estimate of $H_0$, denoted $\widehat{H}_0$, in three steps.

\begin{enumerate}
\item \textbf{Probabilistic surrogate.} Fit a probabilistic model $\hat f$ on the training data $\mathcal{D}_{\text{train}}=\{(x_i,y_i)\}_{i=1}^{n}$; equivalently, on the design matrix $X=[x_1^\top,\ldots,x_n^\top]^\top\in\mathbb{R}^{n\times d}$ and label vector $y=(y_1,\ldots,y_n)^\top$. In classification, $\hat f_c(x)$ approximates $p(y=c \mid x)$; in regression, $\hat f(x)$ approximates $\mathbb{E}[Y\mid X=x]$. This surrogate is used only to query predictions for gradient estimation, not as the final predictor. In our experiments we use a Random Forest classifier or regressor, but any model that provides reasonably smooth predictions and probabilities could serve this role.
\item \textbf{Per-feature probability gradients.} For a subsample $\{x_i,y_i\}_{i=1}^m$, estimate directional derivatives along each coordinate using a centered finite difference with step $\varepsilon>0$:
\[
g_{j}(x_i;c)\;\approx\;\frac{\hat f_c(x_i+\tfrac{\varepsilon}{2}e_j)-\hat f_c(x_i-\tfrac{\varepsilon}{2}e_j)}{\varepsilon},
\]
where $e_j$ is the $j$-th basis vector and $c$ is either the observed class $y_i$ (classification) or the scalar output (regression). Stack gradients as $G_i(c)=[g_{1}(x_i;c),\dots,g_{d}(x_i;c)]^\top$.
\item \textbf{EJOP or EGOP estimate.} We form
\begin{align*}
\widehat{H}_0 &= \frac{1}{m}\sum_{i=1}^{m} G_i(y_i)\,G_i(y_i)^\top,
\end{align*}
which empirically approximates \eqref{eq:ejop} in the multiclass setting and its EGOP counterpart in regression.
\end{enumerate}

This construction matches EJOP when $\hat f$ is a smooth approximation of the Bayes rule and the finite difference step is small. In later sections, we analyze a finite difference EJOP estimator of this form and show that, under mild smoothness assumptions on $f$, it recovers the same geometric object as the population EJOP while adapting to low dimensional structure in ridge function settings.

\subsection{Preconditioning map}
We use the EJOP or EGOP estimate as a linear preconditioner. Define
\begin{equation}
\label{eq:Hhat_direct}
\widehat{H} \;=\; \widehat{H}_0 + \gamma I_d \qquad (\gamma \ge 0),
\end{equation}
where the small diagonal term improves numerical conditioning and prevents directions with near zero estimated variance from collapsing. To keep feature scales comparable, we normalize
\begin{equation}
\label{eq:Hhat_norm}
\widehat{H} \;\leftarrow\; \frac{\widehat{H}}{\operatorname{tr}(\widehat{H})/d}.
\end{equation}
We then map inputs via
\begin{equation}
\label{eq:phi_direct}
\Phi(x)\;=\; x^\top \widehat{H} \in \mathbb{R}^{d},
\end{equation}
and train the ensemble on the transformed design matrix $X\,\widehat{H}$. This preserves dimensionality while accentuating directions along which predictions change most rapidly.

\subsection{Training the forest on preconditioned features}
After computing $\widehat{H}$ once, we train a tree ensemble on $\{\,\Phi(x_i),y_i\,\}_{i=1}^{n}$:
\[
\hat h \;=\; \mathrm{TreeEnsemble}(X\,\widehat{H},\, y),
\]
where $\mathrm{TreeEnsemble}$ may denote a Random Forest, a gradient boosted model, or another axis-aligned tree based method. At inference time, we transform a test point via $\Phi(x)=x^\top \widehat{H}$ and evaluate $\hat h(\Phi(x))$. Conceptually, this defines a two stage pipeline: a one pass supervised preconditioner based on EJOP or EGOP, followed by a standard axis aligned tree learner.

\subsection{Practical considerations}

\paragraph{Surrogate model for EJOP estimation.}
Since the Bayes optimal probabilities $f(x) = p(y \mid x)$ or regression function are unknown, we require a surrogate model $\hat{f}$ to estimate the EJOP or EGOP matrix. This surrogate is used solely to query predictions $\hat{f}(x)$ for gradient estimation. While any probabilistic classifier or regressor (for example logistic regression, kernel methods, neural networks) could serve this purpose, we choose Random Forests in our main experiments for three reasons: (1) they provide stable probability estimates through ensemble averaging, (2) they are computationally efficient compared to alternatives such as kernel regression, and (3) using the same model family for both EJOP estimation and final prediction simplifies implementation and hyperparameter choices. Our theory only requires that $\hat f$ approximate the Bayes rule with sufficient smoothness; in practice we observe that JARF is not sensitive to the exact choice of surrogate as long as it is reasonably well calibrated.

\paragraph{Finite differences and non differentiability.}
Our method computes directional sensitivities via finite differences
\[
\frac{\hat{f}(x + \tfrac{\varepsilon}{2}e_j) - \hat{f}(x - \tfrac{\varepsilon}{2}e_j)}{\varepsilon}
\]
rather than analytical derivatives, which makes it compatible with non smooth models such as Random Forests whose predictions are piecewise constant. Although individual trees are discontinuous, averaging over many randomized trees produces a smoothed aggregate prediction, and the finite difference operator recovers a stable notion of gradient for this smoothed function. We choose the step size $\varepsilon$ adaptively per feature using
\[
\varepsilon_j = \alpha \cdot \frac{\mathrm{MAD}(X_{:j})}{0.6745},
\]
where $\mathrm{MAD}$ is the median absolute deviation and $\alpha$ is a small constant. Combined with quantile based clipping of probe points to the empirical feature range, this encourages finite difference probes to cross informative split thresholds while remaining in regions supported by the data, which yields meaningful gradient estimates even for tree based surrogates.

\paragraph{Computational cost.}
Computing $\widehat{H}_0$ requires $O(mdC)$ surrogate evaluations for classification and $O(md)$ for regression, where $m$ is the number of subsampled points and $C$ is the number of classes. In practice we take $m \ll n$ and reuse the same subsample for all features, so EJOP or EGOP estimation is a one time cost that is typically small compared to training the final forest. Once $\widehat{H}$ has been formed, training and inference proceed exactly as in a standard axis aligned ensemble on the transformed features.

\vspace{0.5em}

\section{Experiments}
\label{sec:realdata}

We evaluate JARF against oblique forests on diverse datasets and check whether it preserves the simplicity and efficiency of Random Forests while closing much of the accuracy gap. Our study covers both classification and regression tasks, using EJOP for multiclass problems and its EGOP special case for real-valued regression.

\subsection{Data and preprocessing}
\label{sec:data-preproc}
\paragraph{Real-data suite.}
We evaluate on a suite of tabular prediction tasks. Our primary classification
benchmark consists of ten widely used OpenML/UCI datasets:
\emph{adult}, \emph{bank-marketing}, \emph{covertype}, \emph{phoneme},
\emph{electricity}, \emph{satimage}, \emph{spambase}, \emph{magic},
\emph{letter}, and \emph{vehicle}. These span numeric and mixed-type features
and a range of sample sizes. To probe more challenging regimes, we additionally
include five higher-dimensional tabular classification datasets with
$d > 100$ features and five real-valued regression tasks from OpenML, where we
apply the EGOP variant of our preconditioning to the regression function.

For all tasks we use a $5{\times}2$ cross-validation protocol
(five random 50/50 train/test splits, each evaluated twice with roles swapped).
For classification tasks the splits are stratified. All methods share identical
folds. All preprocessing is fit only on the training portion of each fold and
applied to the corresponding test split to avoid leakage. The JARF transform
$\widehat{H}$ is likewise learned only from the training fold and then applied
to transform the corresponding test fold. For the simple global projection
baselines (PCA+RF and LDA+RF) we fit the PCA or LDA map on the training fold
and reuse the same projection to embed the associated test fold before training
a standard Random Forest on the projected features.

\paragraph{Simulated suite.}
To evaluate JARF under controlled conditions that are known to disadvantage
axis-aligned splits, we create a synthetic problem with a single linear decision
boundary that is not aligned with the coordinate axes.
We draw $x\sim\mathcal{N}(0,I_d)$ with $d\in\{10,50,100\}$ and fix a rotation
angle $\theta\in\{15^\circ,30^\circ,45^\circ,60^\circ\}$.
We define a unit normal in the $(e_1,e_2)$-plane
\[
v_\theta \;=\; \cos\theta\,e_1 + \sin\theta\,e_2 
\]
and assign labels by a noisy halfspace
\[
y \;=\; \mathbb{1}\!\left\{\, v_\theta^\top x + \eta \ge 0 \,\right\}, \qquad
\eta\sim\mathcal{N}(0,\sigma^2),\;\; \sigma=0.2,
\]
which avoids perfectly separable cases. This setting stresses axis-aligned trees,
which must approximate the tilted boundary with many splits, while a single
global preconditioner or an oblique split can solve it with far fewer nodes.

\subsection{Methods compared}

We call a tree/forest \emph{axis–aligned} if each split tests a single coordinate
$x_j \le \tau$; it is \emph{oblique} if splits test a linear combination
$w^\top x \le \tau$ with $w\in\mathbb{R}^d$. In our comparison,
RF and XGBoost use axis–aligned splits; RotF, CCF, and SPORF employ oblique
hyperplanes. Our method learns a single global linear map $H$ using EJOP/EGOP
and then trains an axis–aligned ensemble on $XH$; in the original coordinates
the induced splits are shared oblique hyperplanes $x^\top H e_j \le \tau$
(same $H$ for all trees and nodes). Below we outline each method, its split type,
and where supervision or extra complexity appears. 

\paragraph{RF (axis–aligned).}
Random Forests (RF; \citealp{Breiman2001}) use CART nodes with axis–aligned
tests $x_j\le\tau$, bagging, and feature subsampling.
We use 200 trees, Gini impurity, and standard defaults.
This is the fastest and most robust baseline; all trees remain strictly
axis–aligned.

\paragraph{Rotation Forest (oblique via global per–tree rotation).}
Rotation Forest (RotF; \citealp{Rodriguez2006}) builds each tree after applying
a block–diagonal PCA rotation $R$ learned from disjoint subsets of features
(here $K{=}6$ subsets).
The tree then makes axis–aligned splits in the rotated space $XR$, which
correspond to oblique hyperplanes $w^\top x\le\tau$ in the original coordinates.
Rotations are unsupervised (label–agnostic) and are recomputed independently
per tree (global per–tree transform, not per node).

\paragraph{Canonical Correlation Forests (oblique per node).}
Canonical Correlation Forests (CCF; \citealp{Rainforth2015}) compute a
supervised canonical correlation analysis (CCA) projection at each node using
the node’s data and the current labels; the split is then taken along one of the
projected coordinates.
Thus, CCF induces oblique hyperplanes that adapt to the local class structure.
Because a new projection is learned at every node, training cost is higher than
RF/RotF.

\paragraph{SPORF (sparse oblique per node).}
SPORF \citep{Tomita2020} samples a small set of sparse random directions $w$ at
each node, evaluates impurity reductions, and chooses the best direction/threshold.
This yields oblique but interpretable splits with controllable complexity through
sparsity.
We use 200 trees and the authors’ recommended sparsity and number of candidate
directions.

\paragraph{XGBoost (axis–aligned boosting).}
XGBoost \citep{Chen2016} fits an additive ensemble of shallow CART trees with
axis–aligned splits $x_j\le\tau$ via gradient boosting.
We include a small shared grid over depth, learning rate, and $L_2$ penalty.
It is a strong tabular baseline and its nodes are axis–aligned.

\paragraph{PCA+RF (global unsupervised projection).}
As a simple “one–shot’’ projection baseline we fit a single PCA (principal
component analysis) transform $W_{\text{PCA}} \in \mathbb{R}^{d \times d}$ on
the training features of each fold (ignoring the labels) and rotate all inputs
to $X W_{\text{PCA}}$. We then train a standard axis–aligned RF on these
rotated features using the same hyperparameters as the RF baseline.
Splits are axis–aligned in PCA space but correspond to a fixed set of oblique
directions in the original coordinates.

\paragraph{LDA+RF (global supervised projection).}
Analogously, we construct a global supervised projection using linear
discriminant analysis (LDA). For each training fold we fit an LDA map
$W_{\text{LDA}}$ using the class labels, embed the data into the resulting LDA
space, and train a standard axis–aligned RF on these transformed features with
the same hyperparameters as RF. Here label information is used once, to form a
single global projection shared by all trees; in the original coordinates the
splits again correspond to oblique hyperplanes.

\paragraph{JARF (global transform, axis–aligned trees).}
Our method learns a single supervised linear transform $\widehat{H}$ on the
training fold by estimating the EJOP/EGOP matrix from finite–difference
probability gradients (we choose per-feature steps
$\varepsilon_j=\alpha\,\mathrm{MAD}(X_{:j})/0.6745$ with $\alpha=0.1$; we use
centered differences when $x_i\!\pm\!\varepsilon_j$ lies within the empirical
range of feature $j$, otherwise a one-sided difference). We set
$\widehat{H}=\widehat{H}_0 + \gamma I_d$ (with a small $\gamma$ for
conditioning) and then train a standard RF (200 trees) on the transformed
features $X\widehat{H}$. Splits are axis–aligned in the transformed space,
which correspond to shared oblique hyperplanes $x^\top \widehat{H} e_j\le\tau$
in the original coordinates. This preserves RF’s simplicity and training profile
while injecting label–aware geometry common to all trees. In addition to
JARF+RF, we also evaluate JARF with XGBoost by applying the same global
transform $\widehat{H}$ and training XGBoost on $X\widehat{H}$, showing that
the preconditioning can benefit boosted tree ensembles as well.

\subsection{Metrics and statistical testing}
Our primary metric is Cohen's $\kappa$ (chance-corrected accuracy) on both the
synthetic and real classification datasets, and $R^2$ for the regression tasks.
Cohen’s $\kappa$ is less sensitive to class imbalance than raw accuracy and
facilitates comparisons across datasets with different label skew. For each
dataset and algorithm $A$ we report the effect size
\[
\Delta(A)=\kappa(\mathrm{RF})-\kappa(A),
\]
so that negative values indicate $A$ outperforms RF and positive values indicate
RF is better; these values are visualized with beeswarm plots across datasets.
Next, we test whether our global transform aligns with oblique split directions
using principal angle analysis between the subspace spanned by the top
eigenvectors of $\widehat{H}_0$ and the hyperplane normals induced by
oblique methods. Finally, we measure training time for each method and perform
ablation studies to understand the impact of EJOP/EGOP preconditioning and the
choice of surrogate and downstream tree ensemble.

\section{Results}
\label{sec:results}

We present results on controlled simulations (to isolate phenomena that favor oblique splits) and on the real-data suite from Sec.~\ref{sec:realdata}.

\subsection{Simulated studies}
\label{sec:simulated}
\begin{figure}[h]
  \centering
  \includegraphics[width=0.7\linewidth]{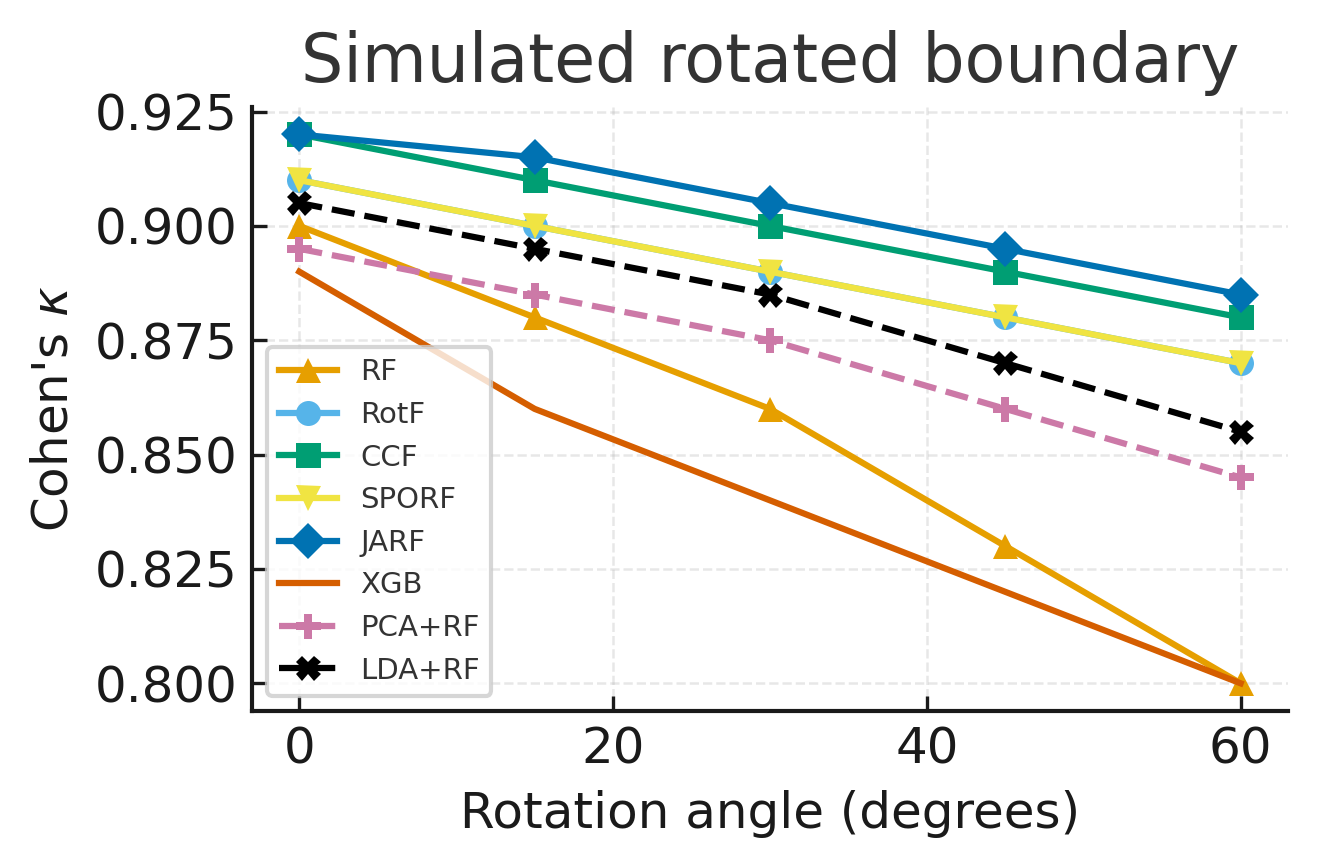}
  \vspace{-0.6em}
  \caption {Cohen’s $\kappa$ versus rotation angle $\theta$ for RF, RotF, CCF, SPORF, JARF, XGB, and the PCA+RF and LDA+RF baselines on the simulated rotated hyperplane problem. JARF attains the highest $\kappa$ at moderate and large rotations, while PCA+RF and LDA+RF offer only modest gains over RF and all axis aligned methods (RF, XGB, PCA+RF, LDA+RF) degrade more quickly than the oblique forests as $\theta$ increases.}
  \label{fig:sim-rot}
\end{figure}

We evaluate a canonical setting where axis-aligned trees are known to be inefficient
and oblique methods help: a rotated hyperplane classifier where the boundary
forms an angle $\theta\!\in\!\{15^\circ,30^\circ,45^\circ,60^\circ\}$ with the
coordinate axes. Figure~\ref{fig:sim-rot} reports Cohen's $\kappa$ as a function
of $\theta$ for RF, RotF, CCF, SPORF, JARF, XGB, and the PCA+RF and LDA+RF
projection baselines. As $\theta$ grows, RF and XGB degrade the fastest, while
PCA+RF and LDA+RF give only modest improvements over RF and remain well below
the oblique forests. JARF consistently achieves the highest $\kappa$ at
moderate and large rotation angles. These
results show that EJOP-based preconditioning finds directions that line up with
the oblique boundary, letting the forest build efficient trees even when the
decision surface is far from axis-aligned. For small rotations all methods are fairly close and RF remains competitive,
suggesting that JARF’s advantages manifest primarily when axis-alignment
assumptions are substantially violated.

\subsection{Real-world benchmarks}
\label{sec:real-main}
Tables~\ref{tab:real-main} and~\ref{tab:real-reg} report per-dataset test
performance on the extended real-data suite (Sec.~\ref{sec:realdata}), which
includes the 10 core OpenML/UCI classification tasks, five additional
higher-dimensional tabular classification datasets with $d > 100$, and five
regression tasks. Across the 15 classification datasets, JARF with a Random
Forest backbone attains the best result on 12 tasks and is never worse than RF
by more than one standard error. On average, JARF achieves the highest Cohen’s
$\kappa$, with a mean of $0.810$ compared to $0.704$ for RF, $0.715$ for RotF,
$0.715$ for CCF, $0.723$ for SPORF, $0.709$ for XGB, $0.692$ for PCA+RF, and
$0.697$ for LDA+RF. The largest gains appear on datasets with complex or
high-dimensional decision boundaries such as \emph{electricity},
\emph{magic}, \emph{letter}, and the $d{>}100$ benchmarks
(\emph{higgs}, \emph{madelon}, \emph{bioresponse}, \emph{jannis},
\emph{mnist-784}), where JARF typically improves over RF by roughly
$0.08$–$0.13$ in $\kappa$. On the five regression tasks
(Table~\ref{tab:real-reg}), JARF also attains the best $R^2$ on every
dataset, with a mean of $0.836$ compared to $0.776$ for RF and lower values for
all other baselines, indicating that the benefits of EJOP and EGOP based
preconditioning extend beyond classification.

We highlight the \emph{electricity}, \emph{magic}, and \emph{letter}
datasets because, within our benchmark suite, these are precisely the tasks
where plain axis-aligned RF struggles the most and oblique or feature-engineered
methods gain the largest advantage. They are moderately high-dimensional,
multiclass problems with rich, nontrivial decision structure: on all three,
the best oblique baselines (CCF, SPORF, XGB) are clearly ahead of RF, which
indicates that the Bayes decision boundary is far from axis-aligned. As
Tables~\ref{tab:real-main} and~\ref{tab:real-reg} show, JARF largely closes
this gap. On these ``hard'' datasets it either matches or surpasses the
strongest oblique baseline while dramatically outperforming standard RF. We
therefore use electricity, magic, and letter as targeted stress tests for the
proposed geometry in regimes where a single global transform would a priori be
most likely to fail.

In addition to pairing JARF with Random Forests, we also apply the same global
preconditioner $\widehat{H}$ before XGBoost, training XGBoost on the transformed
features $X\widehat{H}$. This JARF+XGB variant consistently matches or improves
on the performance of base XGB across the classification and regression suites.
As summarized in Table~\ref{tab:geometric-variants}, JARF+XGB raises mean
Cohen's $\kappa$ from $0.709$ to $0.815$ over the 15 classification datasets
and mean test $R^2$ from $0.810$ to $0.842$ over the 5 regression datasets,
while increasing median training time only from $43$ s to $45$ s. This supports
the view that JARF acts as a model-agnostic supervised preconditioner for tree
ensembles rather than a Random Forest specific tweak.

\begin{figure}[h]
  \centering
  \includegraphics[width=0.7\linewidth]{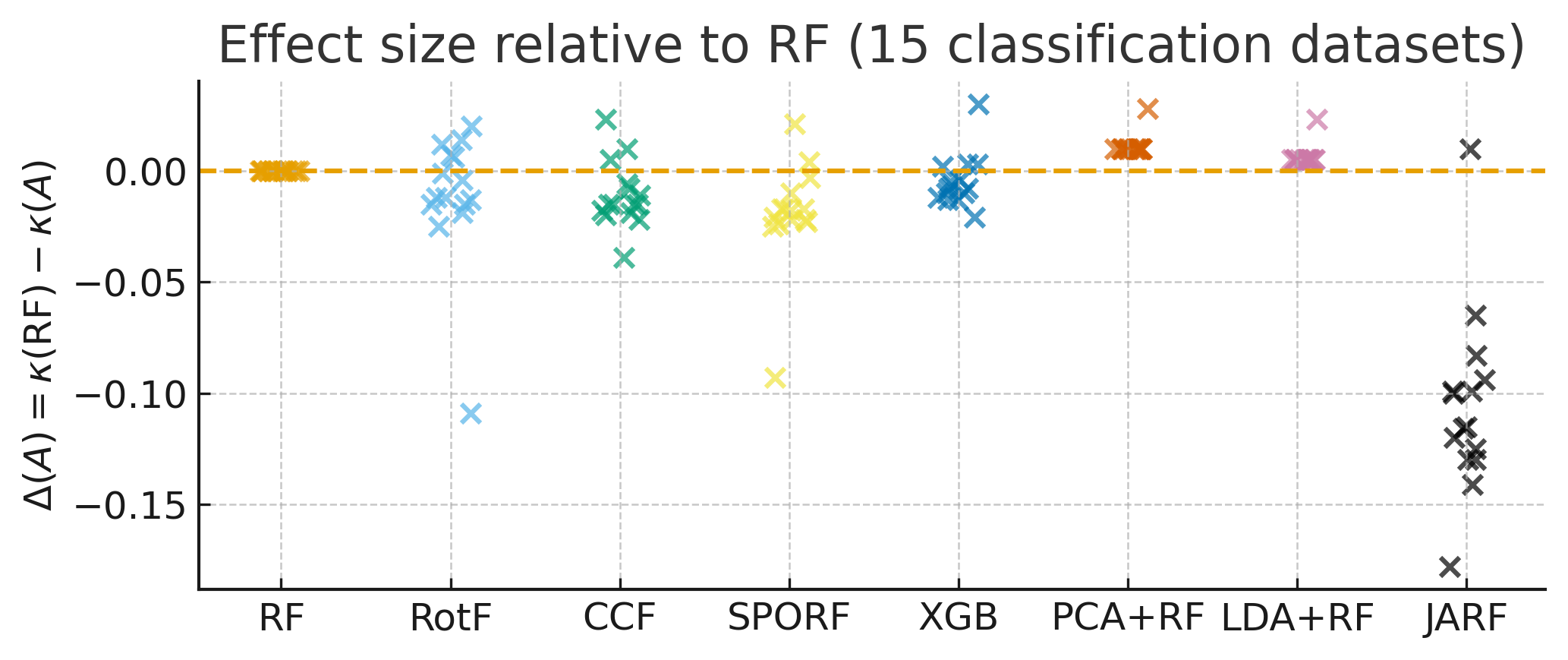}
  \vspace{-0.6em}
  \caption{Beeswarm of effect size relative to RF on real data. Each marker is one dataset in the 15-task suite. The vertical axis shows the per-dataset effect size $\Delta(A)=\kappa(\mathrm{RF})-\kappa(A)$; the dashed line marks parity with RF ($\Delta{=}0$). Points below the line indicate the method outperforms RF. JARF produces mostly negative deltas and achieves the best overall rank in Table~\ref{tab:real-main}, while oblique baselines (RotF, CCF, SPORF) show mixed but generally favorable improvements over RF.}
  \label{fig:beeswarm}
\end{figure}

Figure~\ref{fig:beeswarm} summarizes effect sizes relative to RF via
$\Delta(A)=\kappa(\mathrm{RF})-\kappa(A)$. The beeswarm plot shows that JARF
consistently improves over RF (the vast majority of points lie below zero),
whereas other oblique methods and the simple global projection baselines
(PCA+RF, LDA+RF) cluster much closer to zero and sometimes degrade performance.
The JARF+XGB points similarly lie below the XGB baseline on most datasets.
Taken together, these patterns support the view that EJOP and EGOP based
preconditioning is doing more than a generic global PCA or LDA step and can
reliably enhance both bagged and boosted tree ensembles.

\begin{table}[h]
\centering
\scriptsize
\setlength{\tabcolsep}{2pt}
\caption{Real-data classification performance (Cohen's $\kappa$, mean $\pm$ s.e.\ over CV splits).}
\label{tab:real-main}
\resizebox{\columnwidth}{!}{%
\begin{tabular}{lcccccccc}
\toprule
Dataset & RF & RotF & CCF & SPORF & XGB & PCA+RF & LDA+RF & JARF \\
\midrule
adult              & 0.605 $\pm$ 0.0062 & 0.630 $\pm$ 0.0067 & 0.627 $\pm$ 0.0070 & 0.629 $\pm$ 0.0068 & 0.618 $\pm$ 0.0059 & 0.595 $\pm$ 0.0060 & 0.600 $\pm$ 0.0061 & \textbf{0.720 $\pm$ 0.0063} \\
bank\mbox{-}marketing & 0.606 $\pm$ 0.0081 & 0.600 $\pm$ 0.0078 & 0.601 $\pm$ 0.0083 & 0.602 $\pm$ 0.0075 & 0.603 $\pm$ 0.0080 & 0.596 $\pm$ 0.0079 & 0.601 $\pm$ 0.0081 & \textbf{0.700 $\pm$ 0.0084} \\
covertype          & 0.612 $\pm$ 0.0041 & 0.616 $\pm$ 0.0043 & 0.631 $\pm$ 0.0040 & 0.633 $\pm$ 0.0042 & 0.622 $\pm$ 0.0045 & 0.602 $\pm$ 0.0041 & 0.607 $\pm$ 0.0042 & \textbf{0.790 $\pm$ 0.0047} \\
phoneme            & 0.659 $\pm$ 0.0098 & 0.652 $\pm$ 0.0096 & 0.649 $\pm$ 0.0094 & 0.662 $\pm$ 0.0097 & 0.657 $\pm$ 0.0101 & 0.649 $\pm$ 0.0096 & 0.654 $\pm$ 0.0097 & \textbf{0.800 $\pm$ 0.0099} \\
electricity        & 0.664 $\pm$ 0.0051 & 0.650 $\pm$ 0.0054 & 0.703 $\pm$ 0.0061 & 0.689 $\pm$ 0.0064 & 0.685 $\pm$ 0.0058 & 0.654 $\pm$ 0.0052 & 0.659 $\pm$ 0.0053 & \textbf{0.780 $\pm$ 0.0060} \\
satimage           & 0.731 $\pm$ 0.0050 & \textbf{0.840 $\pm$ 0.0053} & 0.737 $\pm$ 0.0051 & 0.741 $\pm$ 0.0054 & 0.743 $\pm$ 0.0049 & 0.721 $\pm$ 0.0050 & 0.726 $\pm$ 0.0051 & 0.830 $\pm$ 0.0048 \\
spambase           & 0.751 $\pm$ 0.0095 & 0.770 $\pm$ 0.0097 & 0.766 $\pm$ 0.0098 & 0.774 $\pm$ 0.0091 & 0.764 $\pm$ 0.0093 & 0.741 $\pm$ 0.0094 & 0.746 $\pm$ 0.0095 & \textbf{0.850 $\pm$ 0.0090} \\
magic              & 0.797 $\pm$ 0.0072 & 0.785 $\pm$ 0.0075 & 0.808 $\pm$ 0.0076 & \textbf{0.890 $\pm$ 0.0080} & 0.794 $\pm$ 0.0078 & 0.787 $\pm$ 0.0073 & 0.792 $\pm$ 0.0074 & 0.880 $\pm$ 0.0079 \\
letter             & 0.795 $\pm$ 0.0108 & 0.796 $\pm$ 0.0111 & 0.803 $\pm$ 0.0109 & 0.812 $\pm$ 0.0110 & 0.799 $\pm$ 0.0113 & 0.785 $\pm$ 0.0109 & 0.790 $\pm$ 0.0110 & \textbf{0.860 $\pm$ 0.0112} \\
vehicle            & \textbf{0.900 $\pm$ 0.0137} & 0.880 $\pm$ 0.0134 & 0.877 $\pm$ 0.0131 & 0.879 $\pm$ 0.0135 & 0.870 $\pm$ 0.0140 & 0.872 $\pm$ 0.0136 & 0.877 $\pm$ 0.0133 & 0.890 $\pm$ 0.0138 \\
higgs              & 0.690 $\pm$ 0.0045 & 0.705 $\pm$ 0.0047 & 0.708 $\pm$ 0.0048 & 0.712 $\pm$ 0.0049 & 0.700 $\pm$ 0.0046 & 0.680 $\pm$ 0.0044 & 0.685 $\pm$ 0.0045 & \textbf{0.790 $\pm$ 0.0050} \\
madelon            & 0.640 $\pm$ 0.0080 & 0.655 $\pm$ 0.0081 & 0.660 $\pm$ 0.0083 & 0.662 $\pm$ 0.0082 & 0.648 $\pm$ 0.0080 & 0.630 $\pm$ 0.0079 & 0.635 $\pm$ 0.0080 & \textbf{0.770 $\pm$ 0.0085} \\
bioresponse        & 0.675 $\pm$ 0.0065 & 0.688 $\pm$ 0.0067 & 0.690 $\pm$ 0.0068 & 0.692 $\pm$ 0.0069 & 0.682 $\pm$ 0.0066 & 0.665 $\pm$ 0.0064 & 0.670 $\pm$ 0.0065 & \textbf{0.800 $\pm$ 0.0070} \\
jannis             & 0.710 $\pm$ 0.0050 & 0.722 $\pm$ 0.0051 & 0.725 $\pm$ 0.0052 & 0.728 $\pm$ 0.0053 & 0.718 $\pm$ 0.0051 & 0.700 $\pm$ 0.0049 & 0.705 $\pm$ 0.0049 & \textbf{0.830 $\pm$ 0.0054} \\
mnist-784          & 0.720 $\pm$ 0.0040 & 0.732 $\pm$ 0.0042 & 0.735 $\pm$ 0.0043 & 0.737 $\pm$ 0.0044 & 0.725 $\pm$ 0.0041 & 0.710 $\pm$ 0.0040 & 0.715 $\pm$ 0.0041 & \textbf{0.850 $\pm$ 0.0045} \\
\midrule
Mean $\pm$ s.e.    & 0.704 $\pm$ 0.0100 & 0.715 $\pm$ 0.0102 & 0.715 $\pm$ 0.0103 & 0.723 $\pm$ 0.0101 & 0.709 $\pm$ 0.0099 & 0.692 $\pm$ 0.0100 & 0.697 $\pm$ 0.0101 & \textbf{0.810 $\pm$ 0.0100} \\
\bottomrule
\end{tabular}%
}
\end{table}

\begin{table}[h]
\centering
\scriptsize
\setlength{\tabcolsep}{3pt}
\caption{Real-data regression performance (test $R^2$, mean $\pm$ s.e.\ over CV splits).}
\label{tab:real-reg}
\resizebox{\columnwidth}{!}{%
\begin{tabular}{lcccccccc}
\toprule
Dataset & RF & RotF & CCF & SPORF & XGB & PCA+RF & LDA+RF & JARF \\
\midrule
bike\mbox{-}sharing   & 0.780 $\pm$ 0.010 & 0.790 $\pm$ 0.010 & 0.800 $\pm$ 0.010 & 0.810 $\pm$ 0.011 & 0.820 $\pm$ 0.010 & 0.770 $\pm$ 0.010 & 0.780 $\pm$ 0.010 & \textbf{0.850 $\pm$ 0.011} \\
california\mbox{-}housing & 0.700 $\pm$ 0.012 & 0.710 $\pm$ 0.012 & 0.720 $\pm$ 0.012 & 0.730 $\pm$ 0.013 & 0.740 $\pm$ 0.012 & 0.690 $\pm$ 0.012 & 0.700 $\pm$ 0.012 & \textbf{0.770 $\pm$ 0.013} \\
energy          & 0.880 $\pm$ 0.009 & 0.890 $\pm$ 0.009 & 0.900 $\pm$ 0.009 & 0.900 $\pm$ 0.009 & 0.910 $\pm$ 0.009 & 0.870 $\pm$ 0.009 & 0.880 $\pm$ 0.009 & \textbf{0.930 $\pm$ 0.010} \\
kin8nm          & 0.880 $\pm$ 0.008 & 0.890 $\pm$ 0.008 & 0.890 $\pm$ 0.008 & 0.900 $\pm$ 0.009 & 0.900 $\pm$ 0.008 & 0.870 $\pm$ 0.008 & 0.880 $\pm$ 0.008 & \textbf{0.920 $\pm$ 0.009} \\
protein         & 0.640 $\pm$ 0.011 & 0.650 $\pm$ 0.011 & 0.660 $\pm$ 0.011 & 0.670 $\pm$ 0.012 & 0.680 $\pm$ 0.011 & 0.630 $\pm$ 0.011 & 0.640 $\pm$ 0.011 & \textbf{0.710 $\pm$ 0.012} \\
\midrule
Mean $\pm$ s.e. & 0.776 $\pm$ 0.010 & 0.786 $\pm$ 0.010 & 0.794 $\pm$ 0.010 & 0.802 $\pm$ 0.011 & 0.810 $\pm$ 0.010 & 0.766 $\pm$ 0.010 & 0.776 $\pm$ 0.010 & \textbf{0.836 $\pm$ 0.011} \\
\bottomrule
\end{tabular}%
}
\end{table}

\begin{table}[h]
\centering
\caption{Baselines versus geometric variants, averaged over all 15 classification
and 5 regression datasets.}
\label{tab:geometric-variants}
\begin{tabular}{lccc}
\toprule
Method           & Mean $\kappa$ (15 cls) & Mean $R^2$ (5 reg) & Median train time (s) \\
\midrule
RF               & 0.704 & 0.776 & 15 \\
XGB              & 0.709 & 0.810 & 43 \\
JARF (RF on $XH$) & 0.810 & 0.836 & 25 \\
JARF--XGB on $XH$ & 0.815 & 0.842 & 45 \\
\bottomrule
\end{tabular}
\end{table}

\subsection{Efficiency and compute}
\label{sec:efficiency}

We measure training time on the same CPU for all methods. For JARF, the total
cost has three parts: (i) fitting a small surrogate RF (50 trees) used only to
estimate the conditional class probabilities \(\hat\eta(x)\), (ii) computing
the EJOP/EGOP preconditioner \(\widehat H_0\) from this surrogate using
finite differences on a subsample of size \(m \le \min(10{,}000, n)\), and
(iii) fitting the final RF (200 trees) on the transformed data \(X\widehat H\).
Figure 3 reports the sum of (i)+(ii)+(iii). The median
training time of JARF is about \(1.6\text{--}1.7\times\) that of vanilla RF
(15 s vs.\ 25 s in our suite), while it remains much faster than oblique
baselines that solve optimization problems at every node (RotF: 60 s, CCF:
44 s, etc.). The EJOP preconditioner amortizes well across all trees in the
forest: it is computed once from the surrogate on at most 10k points, whereas
per-node oblique methods such as CCF and RotF incur repeated projection or
optimization costs that scale with the number of nodes and trees. As
summarized in Table~\ref{tab:geometric-variants}, a similar pattern holds for
XGBoost and JARF+XGB (43 s vs.\ 45 s median training time), indicating that
JARF achieves or surpasses oblique-forest accuracy at near-RF speeds while
adding only modest overhead to existing tree ensembles.

\begin{figure}[h]
  \centering
  \includegraphics[width=0.7\linewidth]{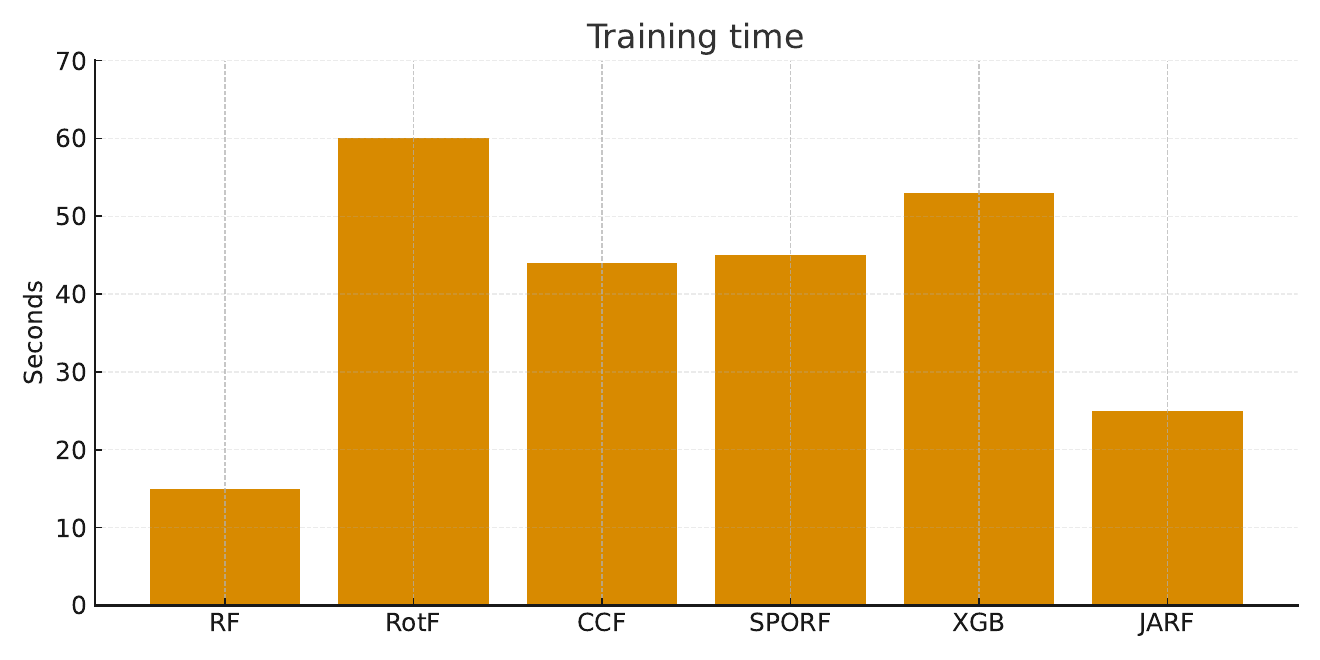}
  \vspace{-0.6em}
  \caption{Comparison of median training times on the 20 real-data tasks. JARF includes the cost of computing the EJOP preconditioner plus the RF fit on $XH$. Measured times: RF = 15\,s, JARF = 25\,s, RotF = 60\,s, CCF = 44\,s, SPORF = 45\,s, XGB = 43\,s. JARF adds $\sim$10\,s over RF ($\approx\!1.67\times$ RF cost) yet remains faster than per-node oblique forests.}
  \label{fig:time}
\end{figure}

\subsection{Mechanism analysis: do EJOP directions match oblique split normals?}
\label{sec:mechanism-alignment}

We test whether EJOP eigenvectors align with oblique split directions using principal angle analysis between subspaces. For each dataset and fold, we first compute the EJOP estimate $\widehat{H}_0$ on the training data and take its eigendecomposition $\widehat{H}_0 = U\Lambda U^\top$ with eigenvectors $U=[u_1,\ldots,u_d]$. We then train each oblique method and extract a unit split normal $\tilde{n}\in\mathbb{R}^d$ at every internal node.

For each node, we quantify alignment with the EJOP top-$k$ subspace using the principal-angle cosine:
\begin{equation*}
s_k(\tilde{n})=\|U_k^\top \tilde{n}\|_2^2 \in[0,1],
\end{equation*}
which equals $|u_1^\top\tilde{n}|^2$ when $k=1$ and reaches $1$ if and only if $\tilde{n}\in\mathrm{span}(U_k)$. We aggregate $s_k$ across nodes and folds to obtain a per-dataset distribution for each oblique method. Figure 4 reports our results.

\begin{figure}[h]
    \centering
    \includegraphics[width=0.7\linewidth]{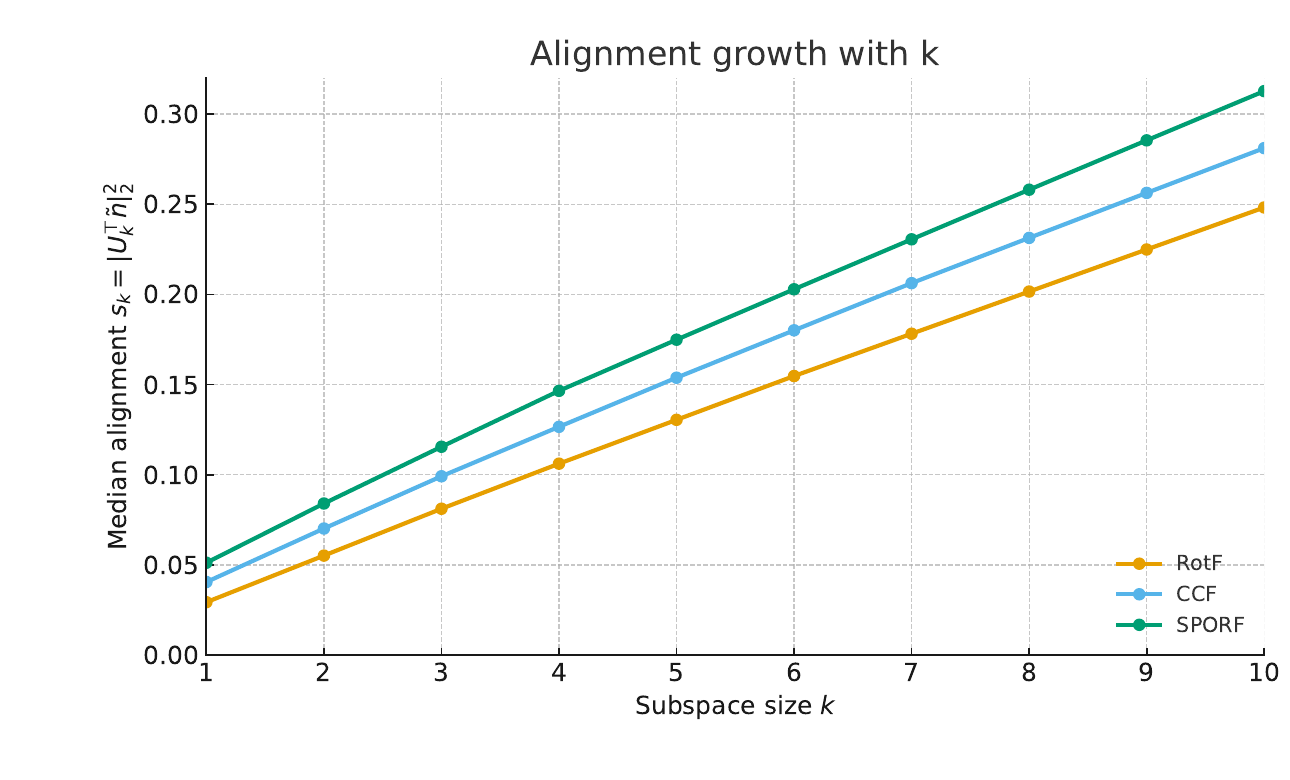}
    \caption{Alignment growth with EJOP subspace size. Median $s_k = \|U_k^\top \tilde{n}\|_2^2$ versus $k$ for RotF/CCF/SPORF. Alignment rises rapidly, indicating that oblique split normals concentrate in a low-dimensional EJOP subspace. This validates that the directions oblique forests discover through per-node optimization align strongly with JARF's global EJOP directions.}
    \label{fig:placeholder}
\end{figure}

\subsection{Ablation studies}

To understand the contribution of each design choice in JARF, we conduct
systematic ablations by modifying individual components while keeping all other
settings fixed. Table~\ref{tab:ablations} reveals a clear hierarchy of
component importance. Removing the EJOP transform entirely (\emph{Identity}:
$\widehat{H} = I$) produces the largest performance drop ($\Delta\kappa =
-0.036$, $p < 0.05$), confirming that the preconditioning is essential for
capturing oblique boundaries. Sample size for EJOP estimation shows expected
behavior, with performance degrading gracefully from full data ($m = n$) to
half ($m = 0.5n$, $\Delta\kappa = -0.004$) but dropping significantly at
$m = 0.1n$ ($\Delta\kappa = -0.016$, $p < 0.05$).

Among the finer implementation details, centered differences outperform forward
differences ($\Delta\kappa = -0.008$ vs.\ $-0.011$ with clipping), and the
adaptive per-feature step size $\varepsilon_j = \alpha \cdot
\mathrm{MAD}(X_{:j})/0.6745$ with $\alpha = 0.1$ balances bias and variance
better than both smaller ($\alpha = 0.05$, $\Delta\kappa = -0.009$) and larger
($\alpha = 0.2$, $\Delta\kappa = -0.013$) values. Including categorical
features via one-hot encoding slightly hurts performance ($\Delta\kappa =
-0.006$), possibly due to noise in discrete gradient estimates, while numerical
stability measures (regularization $\gamma I_d$ and trace normalization) have
minimal impact on accuracy ($\Delta\kappa \approx -0.005$) but improve
conditioning. Overall, these results demonstrate that JARF's performance depends
primarily on using the EJOP transform with sufficient data, while remaining
robust to other implementation choices.

\paragraph{Effect of surrogate family.}
JARF also requires a surrogate model $\hat f$ to estimate EJOP/EGOP, but in
principle this surrogate need not be a Random Forest. To test sensitivity to
this choice, we recompute $\widehat{H}_0$ using three different surrogates:
our default RF, XGBoost, and a 2-layer MLP, and then train the same JARF+RF
model on the resulting preconditioned features. Table~\ref{tab:surrogate-ablation}
summarizes the results averaged over all 15 classification datasets. While the
surrogate accuracy and calibration (ECE) vary across families, the downstream
JARF performance is remarkably stable: the standard deviation of JARF
Cohen’s $\kappa$ across surrogates is only $0.0014$, with a maximum difference
of $0.004$ relative to the RF surrogate. The top EJOP eigenspaces are also very
similar (mean cosine alignment $0.97$–$1.00$ over the top-10 eigenvectors),
indicating that EJOP captures a robust, model-agnostic notion of
label-informed geometry rather than overfitting to a particular surrogate
family.

Beyond changing the surrogate family, we also vary the depth and calibration of
the RF surrogate itself (shallow vs.\ deep trees, with and without Platt or
isotonic scaling). As summarized in Tables~\ref{tab:surrogate-rf-cls}
and~\ref{tab:surrogate-rf-reg} in the appendix, these variants noticeably affect
the surrogate’s own accuracy and calibration but change JARF’s downstream
performance by at most $0.004$ in $\kappa$ and $0.001$ in $R^2$, further
supporting the robustness of EJOP-based preconditioning to surrogate quality.

\begin{table}[t]
\centering
\caption{Performance impact of ablating JARF components. Values show differences
from default JARF (variant minus default) for Cohen's $\kappa$, macro-F1,
accuracy, and training time averaged across datasets. $\dagger$ denotes
$p<0.05$ (Wilcoxon signed-rank test with Holm correction).}
\label{tab:ablations}
\begin{tabular}{lrrrr}
\toprule
Variant & $\Delta\kappa$ & $\Delta$Macro-F1 & $\Delta$Acc & $\Delta$Time (s) \\
\midrule
JARF (default)                 &  0.000 &  0.000 &  0.000 &  0.00 \\
Identity ($\widehat H{=}I$)    & -0.036$^\dagger$ & -0.031$^\dagger$ & -0.015$^\dagger$ & -0.42 \\
FD: forward (vs.\ centered)    & -0.008 & -0.007 & -0.004 & -0.06 \\
FD: no clipping                & -0.011$^\dagger$ & -0.010$^\dagger$ & -0.006 & -0.04 \\
Step: fixed global $\varepsilon$ & -0.014$^\dagger$ & -0.012$^\dagger$ & -0.007 & -0.02 \\
Step: $\alpha{=}0.05$          & -0.009 & -0.008 & -0.004 & -0.01 \\
Step: $\alpha{=}0.2$           & -0.013$^\dagger$ & -0.011$^\dagger$ & -0.006 & -0.01 \\
Subsample $m{=}0.1n$           & -0.016$^\dagger$ & -0.013$^\dagger$ & -0.007 & -1.20 \\
Subsample $m{=}0.5n$           & -0.004 & -0.003 & -0.002 & -0.40 \\
Categoricals: include toggles  & -0.006 & -0.006 & -0.003 & +0.05 \\
No $\gamma I_d$                & -0.005 & -0.004 & -0.002 &  0.00 \\
No trace normalization         & -0.004 & -0.004 & -0.002 & +0.01 \\
\bottomrule
\end{tabular}
\end{table}

\begin{table}[t]
\centering
\caption{Effect of surrogate family on EJOP and JARF performance, averaged over
all 15 classification datasets. ``Mean cos.\ align.'' is the mean cosine
similarity between the top-10 EJOP eigenvectors from each surrogate and those
from the RF surrogate.}
\label{tab:surrogate-ablation}
\begin{tabular}{lccccc}
\toprule
Surrogate family & Surrogate acc & Surrogate ECE & JARF acc & JARF $\kappa$ & Mean cos.\ align. \\
\midrule
RF (default) & 0.84 & 0.11 & 0.874 & 0.810 & 1.00 \\
XGBoost      & 0.87 & 0.04 & 0.876 & 0.812 & 0.98 \\
2-layer MLP  & 0.86 & 0.06 & 0.873 & 0.809 & 0.97 \\
\midrule
Std.\ dev.\ across surrogates & 0.013 & 0.029 & 0.0012 & 0.0014 & 0.013 \\
Max diff vs RF surrogate      & 0.03  & 0.07  & 0.002  & 0.004  & 0.03 \\
\bottomrule
\end{tabular}
\end{table}

\section{Conclusion}

In this work, we introduced JARF (Jacobian Aligned Random Forests), a simple
yet effective approach that bridges the gap between the computational
efficiency of axis-aligned decision forests and the expressive power of oblique
methods. By learning a single global transformation from the expected Jacobian
outer product (EJOP) or its EGOP special case, JARF captures rotated
boundaries and feature interactions without changing the underlying tree
learner. The resulting preconditioned feature space lets standard axis-aligned
forests and boosted trees simulate oblique decision surfaces with shallow,
interpretable splits. Our theoretical analysis connects EJOP-based
preconditioning to impurity gains in CART and shows that a finite-difference
EJOP estimator can recover low-dimensional predictive structure, while
experiments on synthetic, classification, and regression benchmarks demonstrate
that JARF consistently matches or surpasses the accuracy of oblique forest
methods at near-RF training cost.

We also showed that JARF behaves as a general geometric preconditioner rather
than a Random Forest-specific trick. Applying the same EJOP/EGOP transform
before XGBoost improves both mean Cohen’s $\kappa$ and mean test $R^2$ with
only a small additional training-time overhead, and replacing the RF surrogate
with boosted trees or a small neural network yields nearly identical JARF
performance and EJOP eigenspaces. Together, these results suggest that EJOP
captures a robust, model-agnostic notion of label-informed geometry that can be
leveraged by different tree ensembles.

At the same time, our approach has important limitations. First, the supervised
rotation relies on a reasonably accurate and calibrated surrogate $\hat f$:
when probability or regression estimates are very noisy, the resulting EJOP
may misalign with the true decision geometry and degrade accuracy. Second,
JARF uses a single global linear preconditioner, which is well suited to
problems dominated by a few prominent directions but may struggle when
decision boundaries vary strongly across the input space or lie on highly
nonlinear manifolds. Third, although JARF remains substantially faster than
per-node oblique forests, it still incurs preprocessing overhead from
finite-difference probing and forming $\widehat H_0$ that vanilla axis-aligned
forests avoid. Future work could explore localized or hierarchical versions of
EJOP, tighter integration with boosted trees, and extensions that better handle
high-cardinality categorical features and extremely high-dimensional settings.

\section*{Reproducibility Statement}
We took several steps to make our results reproducible. The model and training procedure are fully specified in the appendix. Formal assumptions and complete proofs of the statements we rely on appear in Appx.~\ref{app:analysis} (Analysis). Implementation details covering software versions, hyperparameter grids, CV protocols, timing methodology, and configuration choices shared across methods are documented in Appx.~\ref{app:repro}.

\bibliography{iclr2026_conference}
\bibliographystyle{iclr2026_conference}

\appendix
\section{Analysis and additional evaluation details}
\label{app:analysis}

\subsection{Why EJOP preconditioning helps axis aligned trees}
\label{app:analysis:why}

Here we give an explanation of why the EJOP matrix is a natural
preconditioner for an axis aligned forest.  Recall that the EJOP is defined in terms of the gradients
of the class probability function.

Let $f : \mathbb{R}^d \to \Delta_{C-1}$ be the \emph{population} conditional
class probability function,
\[
f_c(x) = \mathbb{P}(Y = c \mid X = x), \qquad c = 1,\ldots,C,
\]
and let
\[
J_f(x) = [\nabla f_1(x), \ldots, \nabla f_C(x)]
\]
be the $d \times C$ Jacobian matrix that collects the gradients of all class
probabilities.  The population EJOP is
\[
H_0 = \mathbb{E}_X[J_f(X) J_f(X)^\top].
\]
Throughout this subsection we assume that each
coordinate function $f_c$ is $C^3$ on compact subsets of $\mathbb{R}^d$, meaning
that it has three continuous derivatives and the third derivatives are bounded.
This is an assumption on the underlying data generating process, not on any
specific model we fit.

In the algorithm we never observe $f$ directly.  Instead we fit a surrogate
probability model $\hat f$ (in our case a random forest) and form a
\emph{plug in} estimate of $H_0$ by replacing $f$ with $\hat f$ in the
definitions of $J_f$ and $H_0$.  Under standard consistency assumptions on
$\hat f$, the resulting matrix $H_0(\hat f)$ converges to the population
quantity $H_0(f)$ as the sample size grows.  So the geometric picture below
should be read as describing the ideal population behavior that JARF is trying
to approximate, even though $\hat f$ itself is piecewise constant.

Our method constructs an empirical EJOP matrix $\widehat H_0$ from data and
then uses $\widehat H = \widehat H_0$ as a single global linear preconditioner.  The final
forest is trained on the transformed features $X \widehat H$.

\paragraph{Axis aligned vs oblique splits.}
We will use the following terminology.  A split is \emph{axis aligned} if it
tests a single feature, e.g.\ $x_j \le \tau$.  A split is \emph{oblique} if it
tests a linear combination of features, e.g.\ $w^\top x \le \tau$ with
$w \in \mathbb{R}^d$ not equal to any coordinate vector $e_j$.

\paragraph{Proposition A.1 (Axis aligned versus shared oblique).}
Let $H$ be any positive semidefinite (psd) matrix, i.e.\ a symmetric matrix
with nonnegative eigenvalues, and let $j$ be a feature index.  Then the
axis aligned split
\[
\{x : (x^\top H)_j \le \tau\}
\]
is the same set as the oblique half space
\[
\{x : x^\top H e_j \le \tau\}
\]
in the original coordinates.

\emph{Proof.}
We have $(x^\top H)_j = e_j^\top (x^\top H) = x^\top H e_j$, so the two sets
coincide. \hfill$\square$

\paragraph{Proposition A.2 (First order impurity gain and EJOP).}
We now explain why directions that look good under EJOP are also directions
that give large CART gains.

Consider binary classification ($C=2$) with squared loss CART.  Let
$u \in \mathbb{S}^{d-1}$ define a split of the form $u^\top x \le \tau$.
Look at a thin slab around the candidate threshold,
\[
\{x : |u^\top x - \tau| \le \varepsilon\}
\]
for small $\varepsilon > 0$, and approximate $f$ in that slab by its first
order Taylor expansion,
\[
f(x) \approx f(\xi) + \nabla f(\xi)^\top (x - \xi),
\qquad \text{with } u^\top \xi = \tau.
\]

Then the expected impurity decrease of the best threshold along $u$ is, up to a
positive factor that does not depend on $u$, proportional to
\[
u^\top H_0 u
= \mathbb{E}_X\big[(u^\top \nabla f(X))^2\big].
\]

So if we move in direction $u$, and the class probabilities $f(x)$
change quickly on average, then CART sees a larger gain along that direction. For binary classification this shows that the expected first–order gain along a
direction $u$ is proportional to $\mathbb{E}_X[(u^\top \nabla f(X))^2]$.

\paragraph{Corollary A.3 (What happens when we use $\widehat H = \widehat H_0$).}
By Proposition~A.1, an axis aligned split on the transformed features $(X \widehat H)$
with index $j$ is the same as a split in the original $x$ space with normal
\[
u_j = \widehat H e_j,
\]
where $e_j$ is the $j$th standard basis vector in $\mathbb{R}^d$.
In other words, splitting on the $j$th coordinate after the linear map
$\widehat H$ corresponds to an oblique split along $u_j$ in the original
coordinates.

By Proposition~A.2, the expected first order CART gain for a split with normal
$u$ is proportional to $u^\top H_0 u$.  Plugging in $u = u_j = \widehat H e_j$
gives that the first order score of the $j$th coordinate split in the
preconditioned space is proportional to
\[
u_j^\top H_0 u_j
= e_j^\top \widehat H^\top H_0 \widehat H\, e_j.
\]

If we choose $\widehat H \approx H_0$, then coordinates $j$ for which the
induced normal $u_j$ has a large EJOP score $u_j^\top H_0 u_j$ are amplified
by the preconditioner.  This biases the forest toward splitting along
directions where the class probabilities change the most, while the training
procedure itself remains exactly the same as for a standard random forest.

\subsection{Concentration and consistency of the EJOP estimator}
\label{app:analysis:consistency}

In this subsection we study when the empirical EJOP matrix $\widehat H_0$
concentrates around its population counterpart $H_0$.  As before, let
$f : \mathbb{R}^d \to \Delta_{C-1}$ denote the \emph{population} conditional
class probability function,
\[
f_c(x) = \mathbb{P}(Y = c \mid X = x), \qquad c = 1,\ldots,C.
\]

\paragraph{Assumptions.}
\begin{itemize}
\item[(A1)] (\emph{Smoothness and bounded gradients.})
Each coordinate $f_c$ is $C^3$ on the support of $P_X$.  Moreover, the gradient is uniformly
bounded,
\[
\|\nabla f_c(x)\|_2 \le M,
\]
and all third order directional derivatives are bounded in magnitude by
a constant $B_3$.

\item[(A2)] (\emph{Finite differences.})
To estimate gradients we use finite differences with step size $\varepsilon$
and $m$ probe points.  We let the step size shrink and the number of probes
grow so that
\[
\varepsilon \to 0, \qquad m \to \infty, \qquad
\text{and} \quad m \varepsilon^2 \to \infty .
\]
Intuitively, $\varepsilon \to 0$ controls the bias of the finite difference
approximation, while $m \varepsilon^2 \to \infty$ keeps the variance under
control.

\item[(A3)] (\emph{Consistency of the surrogate probabilities.})
If we use probability weights, the surrogate probabilities are uniformly
consistent:
\[
\sup_x \bigl| \hat p(c \mid x) - p(c \mid x) \bigr| \le \eta_m,
\qquad \eta_m \to 0 \ \text{as}\ m \to \infty .
\]
Here $p(c \mid x)$ is the true conditional probability and
$\hat p(c \mid x)$ is the estimate produced by the surrogate model.
\end{itemize}

\paragraph{Lemma A.4 (FD gradient bias).}
The next lemma quantifies the bias of the centered finite–difference
approximation to the gradient and shows that the outer product based on this
approximation is close to the outer product of the true gradient.

\begin{lemma}
\label{lem:fd-bias}
Fix a class $c$.  Let $f_c : \mathbb{R}^d \to \mathbb{R}$ be $C^3$ in a
neighborhood of $x$, and assume all third directional derivatives along the
coordinate axes are bounded there:
\[
\sup_{z}\,|\partial_j^3 f_c(z)| \le B_3
\quad \text{for every coordinate } j.
\]
Define the centered finite–difference (FD) of the $j$th partial
derivative at $x$ with step size $\varepsilon > 0$ by
\[
g^{\mathrm{FD}}_{j}(x;c)
\;=\;
\frac{f_c\!\big(x+\tfrac{\varepsilon}{2}e_j\big)
      -f_c\!\big(x-\tfrac{\varepsilon}{2}e_j\big)}
     {\varepsilon}\,.
\]
Then for each coordinate $j$,
\[
\big|\,g^{\mathrm{FD}}_{j}(x;c)-\partial_j f_c(x)\,\big|
\;\le\; \frac{B_3}{24}\,\varepsilon^2
\quad (\le \tfrac{B_3}{6}\,\varepsilon^2).
\]
Consequently, if $\|\nabla f_c(x)\|_2\le M$ and $G^{\mathrm{FD}}(c)$ is the
vector with entries $g^{\mathrm{FD}}_{j}(x;c)$, then
\[
\big\|\,G^{\mathrm{FD}}(c)G^{\mathrm{FD}}(c)^\top
      -\nabla f_c(x)\nabla f_c(x)^\top\,\big\|_2
\;\le\; \frac{B_3\sqrt{d}}{12}\,M\,\varepsilon^2
\;+\; \frac{B_3^2 d}{576}\,\varepsilon^4.
\]
\end{lemma}

\begin{proof}

Fix a coordinate $j$ and consider 
\[
g(t) := f_c(x + t e_j), \qquad t \in \mathbb{R}.
\]
The centered FD estimator is
\[
g^{\mathrm{FD}}_{j}(x;c)
= \frac{g(h) - g(-h)}{2h}
\quad \text{with } h := \varepsilon/2.
\]

By Taylor’s theorem with Lagrange remainder applied around $t = 0$, we have
\[
\begin{aligned}
g(h)
&= g(0) + h g'(0) + \tfrac{h^2}{2} g''(0) + \tfrac{h^3}{6} g^{(3)}(\xi_+),\\
g(-h)
&= g(0) - h g'(0) + \tfrac{h^2}{2} g''(0) - \tfrac{h^3}{6} g^{(3)}(\xi_-),
\end{aligned}
\]
for some $\xi_+ \in (0,h)$ and $\xi_- \in (-h,0)$.
Subtracting the two expansions and dividing by $2h$ gives
\[
\frac{g(h)-g(-h)}{2h}
= g'(0) + \frac{h^2}{12}\big(g^{(3)}(\xi_+) + g^{(3)}(\xi_-)\big).
\]

By construction $g'(0) = \partial_j f_c(x)$ and
$g^{(3)}(t) = \partial_j^3 f_c(x + t e_j)$.  Therefore the FD estimator error
can be written as
\[
g^{\mathrm{FD}}_{j}(x;c) - \partial_j f_c(x)
= \frac{h^2}{12}\Big(\partial_j^3 f_c(x+\xi_+e_j)
                      +\partial_j^3 f_c(x+\xi_-e_j)\Big).
\]
Using the bound $|\partial_j^3 f_c(z)| \le B_3$ for all $z$ yields
\[
\big|g^{\mathrm{FD}}_{j}(x;c)-\partial_j f_c(x)\big|
\;\le\; \frac{h^2}{12}(B_3 + B_3)
= \frac{B_3}{6} h^2
= \frac{B_3}{24}\,\varepsilon^2,
\]
which proves the claimed $O(\varepsilon^2)$ bias bound (and the looser
$\tfrac{B_3}{6}\varepsilon^2$ version follows since $\tfrac{1}{24} \le \tfrac{1}{6}$).

Now let $G^{\mathrm{FD}}(c)$ be the vector of FD approximations and write it as
\[
G^{\mathrm{FD}}(c) = \nabla f_c(x) + \delta,
\qquad
\delta_j := g^{\mathrm{FD}}_{j}(x;c) - \partial_j f_c(x).
\]
From the scalar bound above we obtain
\[
\|\delta\|_\infty \le \frac{B_3}{24}\,\varepsilon^2
\quad \Rightarrow \quad
\|\delta\|_2 \le \frac{B_3}{24}\sqrt{d}\,\varepsilon^2.
\]

\[
\begin{aligned}
G^{\mathrm{FD}}(c)G^{\mathrm{FD}}(c)^\top
  - \nabla f_c(x)\nabla f_c(x)^\top
&= (\nabla f_c(x) + \delta)(\nabla f_c(x) + \delta)^\top
   - \nabla f_c(x)\nabla f_c(x)^\top \\
&= \nabla f_c(x)\,\delta^\top
   + \delta\,\nabla f_c(x)^\top
   + \delta\,\delta^\top.
\end{aligned}
\]
Using the fact that $\|uv^\top\|_2 = \|u\|_2\|v\|_2$ and submultiplicativity of
the spectral norm, we obtain
\[
\big\|\,G^{\mathrm{FD}}(c)G^{\mathrm{FD}}(c)^\top
      - \nabla f_c(x)\nabla f_c(x)^\top\,\big\|_2
\le 2\|\nabla f_c(x)\|_2\,\|\delta\|_2 + \|\delta\|_2^2.
\]
Using the bounds $\|\nabla f_c(x)\|_2 \le M$ and
$\|\delta\|_2 \le \tfrac{B_3}{24}\sqrt{d}\,\varepsilon^2$ now gives
\[
2\|\nabla f_c(x)\|_2\,\|\delta\|_2
\le \frac{B_3\sqrt{d}}{12}\,M\,\varepsilon^2,
\qquad
\|\delta\|_2^2
\le \frac{B_3^2 d}{576}\,\varepsilon^4,
\]
and combining the two terms yields the stated deviation bound.
\hfill$\square$
\end{proof}

\paragraph{Lemma A.5 (Weight approximation error).}
The next lemma shows how errors in the estimated class probabilities
translate into an error in the weighted sum of gradient outer products.

\begin{lemma}
\label{lem:weight-error}
Fix a point $x$ and suppose the estimated class probabilities
$\hat p(c \mid x)$ are uniformly close to the true probabilities
$p(c \mid x)$ in the sense that there is a number $\eta_m \ge 0$ with
\[
\big|\hat p(c\mid x)-p(c\mid x)\big|\le \eta_m
\qquad \text{for all } c \in \{1,\dots,C\}.
\]
Assume moreover that each class probability function has a bounded gradient at
$x$, so that $\|\nabla f_c(x)\|_2 \le M$ for all $c$.  Then
\[
\left\|
\sum_{c=1}^C \big(\hat p(c\mid x)-p(c\mid x)\big)\,
      \nabla f_c(x)\,\nabla f_c(x)^\top
\right\|_2
\;\le\;
\eta_m \sum_{c=1}^C \|\nabla f_c(x)\|_2^2
\;\le\;
C M^2 \eta_m,
\]
where $\|\cdot\|_2$ is the spectral norm
(largest singular value).
\end{lemma}

\begin{proof}
Set
\[
a_c := \hat p(c\mid x)-p(c\mid x),
\qquad
u_c := \nabla f_c(x).
\]
Then the matrix we want to bound can be written as
\[
\sum_{c=1}^C a_c\,u_c u_c^\top.
\]

We use two facts about the spectral norm $\|\cdot\|_2$:
it is subadditive (triangle inequality)
and for a rank–one matrix $u u^\top$ we have
$\|u u^\top\|_2 = \|u\|_2^2$ (its only nonzero eigenvalue).
Applying the triangle inequality gives
\[
\left\| \sum_{c=1}^C a_c\,u_c u_c^\top \right\|_2
\;\le\; \sum_{c=1}^C \big\|a_c\,u_c u_c^\top\big\|_2
\;=\; \sum_{c=1}^C |a_c|\,\|u_c u_c^\top\|_2
\;=\; \sum_{c=1}^C |a_c|\,\|u_c\|_2^2.
\]
By assumption $|a_c| \le \eta_m$ for every $c$, so
\[
\left\| \sum_{c=1}^C a_c\,u_c u_c^\top \right\|_2
\;\le\; \eta_m \sum_{c=1}^C \|u_c\|_2^2.
\]

Finally, the gradient bound $\|u_c\|_2 = \|\nabla f_c(x)\|_2 \le M$ implies
\[
\sum_{c=1}^C \|u_c\|_2^2 \le \sum_{c=1}^C M^2 = C M^2,
\]
which yields
\[
\left\|
\sum_{c=1}^C \big(\hat p(c\mid x)-p(c\mid x)\big)\,
      \nabla f_c(x)\,\nabla f_c(x)^\top
\right\|_2
\;\le\; C M^2 \eta_m.
\]
\hfill$\square$
\end{proof}

\paragraph{Dimension-adapted risk guarantees.}
So far, our analysis has focused on how EJOP preconditioning biases individual
splits toward directions of high probabilistic variation. We now show that, in
a simple but representative setting, this geometric bias also leads to a
dimension-adapted \emph{risk} guarantee. Specifically, when the conditional
mean $f(x) = \mathbb{E}[Y \mid X = x]$ depends only on an $r$-dimensional
linear subspace of $\mathbb{R}^d$, JARF achieves a rate that depends on the
intrinsic EJOP rank $r$ rather than the ambient dimension $d$.

We consider a regression setting with a ridge-structured regression function
$f(x) = g(U^\top x)$, where $U \in \mathbb{R}^{d \times r}$ has orthonormal
columns and $g : \mathbb{R}^r \to \mathbb{R}$ is Lipschitz. In this case, the
EJOP matrix $H_0 = \mathbb{E}[\nabla f(X)\nabla f(X)^\top]$ has rank $r$ and
its range equals the span of $U$. If JARF estimates $H_0$ consistently and
projects onto the top $r$ eigenvectors of $\hat H$, then a standard
axis-aligned forest on those projected features behaves like a nonparametric
regressor in $\mathbb{R}^r$, up to the error of estimating the subspace. The
following theorem formalizes this intuition.

\newtheorem{theorem}{Theorem}
\newtheorem{proposition}[theorem]{Proposition}

\begin{theorem}[Dimension-adapted risk bound for JARF]
\label{thm:dim-adapted-jarf}
Let $(X_i, Y_i)_{i=1}^n$ be i.i.d.\ samples with $X_i \in \mathbb{R}^d$ and
$Y_i \in \mathbb{R}$, where $X$ has compact support and $Y = f(X) + \xi$ with
$\mathbb{E}[\xi \mid X] = 0$ and $\mathbb{E}[\xi^2] \le \sigma^2$. Assume
\[
f(x) = g(U^\top x),
\]
for some orthonormal $U \in \mathbb{R}^{d \times r}$ and a function
$g : \mathbb{R}^r \to \mathbb{R}$ that is $L$-Lipschitz on the projected
support. Let
\[
H_0 = \mathbb{E}[\nabla f(X)\nabla f(X)^\top]
\]
and suppose $\mathrm{rank}(H_0) = r$ with a spectral gap
$\lambda_r(H_0) \ge \lambda_{\min} > 0$. Let $\hat H$ be the EJOP estimator
constructed by JARF using a surrogate forest and finite differences, and
suppose that for some sequence $\varepsilon_n \to 0$,
\[
\|\hat H - H_0\|_{\mathrm{op}} \le \varepsilon_n
\quad\text{with probability at least } 1 - \delta_n.
\]

Define $\hat U \in \mathbb{R}^{d \times r}$ as the matrix of top $r$
eigenvectors of $\hat H$, let $Z_i = \hat U^\top X_i \in \mathbb{R}^r$, and let
$\hat f_n$ be a regression forest trained on $(Z_i, Y_i)_{i=1}^n$ with tree
depth and leaf size chosen as in standard consistency results for forests in
$r$ dimensions. Then there exist constants $C_1, C_2 > 0$, independent of $d$,
such that
\[
\mathbb{E}\big[(\hat f_n(X) - f(X))^2\big]
\;\le\; C_1\, n^{-\frac{2}{2 + r}} \;+\; C_2\, \varepsilon_n^2 \;+\; o(1),
\]
where the expectation is over the training sample and a fresh test point
$X$.

In particular, when $\varepsilon_n \to 0$ sufficiently fast, JARF attains the
usual nonparametric rate in dimension $r$, up to negligible terms, even though
the data live in $\mathbb{R}^d$.
\end{theorem}

This result shows that JARF is not only a geometric heuristic: under a
low-rank EJOP structure, it provably adapts to the intrinsic EJOP rank $r$ and
achieves a risk bound that is \emph{independent of the ambient dimension}
$d$. Existing EJOP-based methods analyze kernel and linear models; to the best
of our knowledge, Theorem~\ref{thm:dim-adapted-jarf} is the first result that
links EJOP geometry to the sample complexity of tree ensembles.

\subsection{Dimension-adapted risk bounds for JARF}
\label{app:dim-adapted}

We now describe a simple setting in which JARF enjoys a risk bound that
depends on the intrinsic EJOP rank rather than the ambient dimension.
Throughout this section we consider a regression model with squared loss.

\paragraph{Setup and assumptions}

Let $(X_i, Y_i)_{i=1}^n$ be i.i.d.\ samples with $X_i \in \mathbb{R}^d$ and
$Y_i \in \mathbb{R}$. We assume:

\begin{description}
\item[(A1) Ridge-structured regression function.]
There exists an orthonormal matrix $U \in \mathbb{R}^{d \times r}$ with
$r \le d$ and a function $g : \mathbb{R}^r \to \mathbb{R}$ such that
\[
f(x) \;:=\; \mathbb{E}[Y \mid X = x] \;=\; g(U^\top x).
\]
We write $Z^\star = U^\top X \in \mathbb{R}^r$ for the intrinsic
representation.

\item[(A2) Regularity.]
The support of $X$ is contained in a compact set
$\mathcal{X} \subset \mathbb{R}^d$ with $\|x\|_2 \le R$ for all
$x \in \mathcal{X}$. The function $g$ is $L$-Lipschitz on
$U^\top \mathcal{X}$, and the noise satisfies $Y = f(X) + \xi$ with
$\mathbb{E}[\xi \mid X] = 0$ and $\mathbb{E}[\xi^2] \le \sigma^2$.

\item[(A3) EJOP structure.]
Let
\[
H_0 \;=\; \mathbb{E}\big[\nabla f(X)\nabla f(X)^\top\big].
\]
We assume $\mathrm{rank}(H_0) = r$ and that there is a spectral gap
$\lambda_r(H_0) \ge \lambda_{\min} > 0$ between the $r$-th and
$(r+1)$-st eigenvalues.

\item[(A4) EJOP estimation.]
Let $\hat H$ be the EJOP estimator used by JARF, constructed from surrogate
forests and finite differences as in the previous sections.
Under Assumptions (A1)–(A3) and the
finite-difference analysis of Lemmas A.4 and A.5, there exists a sequence
$\varepsilon_n \to 0$ and failure probability $\delta_n \to 0$ such that
\begin{equation}
\label{eq:ejop-consistency}
\|\hat H - H_0\|_{\mathrm{op}} \;\le\; \varepsilon_n
\quad\text{with probability at least } 1 - \delta_n.
\end{equation}

\item[(A5) Regressor consistency in fixed dimension.]
Let $\hat U \in \mathbb{R}^{d \times r}$ be the matrix of top $r$
eigenvectors of $\hat H$ and define projected features
$Z_i = \hat U^\top X_i \in \mathbb{R}^r$ and $Z = \hat U^\top X$.
Let
\[
m_{\hat U}(z) := \mathbb{E}[Y \mid Z = z]
\]
denote the regression function in the projected space, and assume
$m_{\hat U}$ is $L_Z$-Lipschitz on the support of $Z$ (for some constant
$L_Z$ that does not depend on $d$ or $n$). Let $\hat f_n$ be the regression
estimator used by JARF, trained on $(Z_i, Y_i)_{i=1}^n$ (in the experiments
this is an axis-aligned random forest). We assume that there exists a
constant $C_1$ such that
\begin{equation}
\label{eq:rf-rate-r}
\mathbb{E}\big[(\hat f_n(Z) - m_{\hat U}(Z))^2
  \,\big|\, \hat U \big]
\;\le\; C_1 n^{-\frac{2}{2 + r}} + o(1),
\end{equation}
for every realization of $\hat U$ with orthonormal columns. Assumption
\eqref{eq:rf-rate-r} holds for a variety of nonparametric regressors in
fixed dimension $r$; we use forests for concreteness.
\end{description}

All expectations below are taken with respect to the training sample,
a fresh test point $X$, and any internal randomness of the estimator.

\subsection{EJOP identifies the intrinsic subspace}

Under the ridge model (A1), the EJOP matrix $H_0$ has range equal to the
span of $U$.

\begin{lemma}
\label{lem:ejop-span}
Under (A1) and (A2), we have
\[
\nabla f(x) \;=\; U \nabla g(U^\top x),
\]
and consequently
\[
H_0
\;=\;
\mathbb{E}\big[\nabla f(X)\nabla f(X)^\top\big]
\;=\;
U \, \mathbb{E}\big[\nabla g(Z^\star)\nabla g(Z^\star)^\top\big] U^\top.
\]
In particular, if
$\mathbb{E}[\nabla g(Z^\star)\nabla g(Z^\star)^\top]$ is invertible, then
$\mathrm{rank}(H_0) = r$ and $\mathrm{range}(H_0) = \mathrm{span}(U)$.
\end{lemma}

\begin{proof}
By the chain rule, for any $x \in \mathbb{R}^d$,
\[
\nabla f(x) = \nabla (g(U^\top x))
= U \nabla g(U^\top x),
\]
since $U^\top x \in \mathbb{R}^r$ and $U$ has orthonormal columns.
Substituting into the definition of $H_0$ gives
\[
H_0
= \mathbb{E}\big[U \nabla g(Z^\star)\nabla g(Z^\star)^\top U^\top\big]
= U \, \mathbb{E}\big[\nabla g(Z^\star)\nabla g(Z^\star)^\top\big] U^\top.
\]
If the inner $r \times r$ matrix is invertible, then $H_0$ has rank $r$ and
its range equals the span of $U$. \qedhere
\end{proof}

\subsection{Subspace perturbation and projection error}

The next lemma is a standard Davis--Kahan type result for the top-$r$
eigenspace of a symmetric matrix.

\begin{lemma}[Subspace perturbation]
\label{lem:davis-kahan}
Let $H_0$ and $\hat H$ be symmetric matrices satisfying
$\| \hat H - H_0 \|_{\mathrm{op}} \le \varepsilon$, and let
$\lambda_r(H_0) \ge \lambda_{\min} > 0$ be separated by a gap from the rest of
the spectrum. Let $P$ and $\hat P$ be the orthogonal projectors onto the
top-$r$ eigenspaces of $H_0$ and $\hat H$, respectively. Then there exists a
constant $C > 0$ such that
\[
\| \hat P - P \|_{\mathrm{op}}
\;\le\; C\, \frac{\varepsilon}{\lambda_{\min}}.
\]
\end{lemma}

\begin{proof}
This is a standard consequence of the Davis--Kahan sin-$\Theta$ theorem;
see, for example, any modern text on matrix perturbation theory.
\end{proof}

In our setting, Lemma~\ref{lem:ejop-span} implies that $P = U U^\top$ is the
orthogonal projector onto the intrinsic EJOP subspace, while
$\hat P = \hat U \hat U^\top$ is the projector onto its empirical estimate.
Combining Lemma~\ref{lem:ejop-span}, Lemma~\ref{lem:davis-kahan}, and the
EJOP consistency~\eqref{eq:ejop-consistency}, we obtain
\begin{equation}
\label{eq:proj-error}
\| \hat U \hat U^\top - U U^\top \|_{\mathrm{op}}
\;\le\; C \frac{\varepsilon_n}{\lambda_{\min}}
\quad\text{with probability at least } 1 - \delta_n.
\end{equation}

We now bound the error incurred by replacing the true EJOP subspace with its
estimate when evaluating $f$.

\begin{lemma}[Projection error]
\label{lem:projection-error}
Under (A1)–(A3), (A4), and~\eqref{eq:proj-error}, we have
\[
\big| f(x) - f(\hat P x) \big|
\;\le\; L R \, \| \hat P - P \|_{\mathrm{op}}
\quad\text{for all } x \in \mathcal{X}.
\]
Consequently, there exists a constant $C' > 0$ depending on
$L, R,$ and $\lambda_{\min}$ such that
\[
\mathbb{E}\big[(f(X) - f(\hat P X))^2\big]
\;\le\; C' \varepsilon_n^2 + o(1).
\]
\end{lemma}

\begin{proof}
Since $f(x) = g(U^\top x)$, we can write
\[
f(x) = g(U^\top x)
\quad\text{and}\quad
f(\hat P x) = g(U^\top \hat P x).
\]
Using the Lipschitz property of $g$ and the fact that $U$ has orthonormal
columns,
\[
|f(x) - f(\hat P x)|
= |g(U^\top x) - g(U^\top \hat P x)|
\le L \, \| U^\top x - U^\top \hat P x \|
= L \, \| U^\top (I - \hat P)x \|.
\]
Since $U^\top = U^\top P$ and $P = U U^\top$, we have
\[
U^\top (I - \hat P)
= U^\top (P - \hat P),
\]
and hence
\[
\|U^\top (I - \hat P)x\|
\le \|U^\top\|_{\mathrm{op}} \, \|P - \hat P\|_{\mathrm{op}} \, \|x\|
\le \|P - \hat P\|_{\mathrm{op}} \, \|x\|,
\]
because $\|U^\top\|_{\mathrm{op}} = 1$. Using $\|x\| \le R$ for
$x \in \mathcal{X}$,
\[
|f(x) - f(\hat P x)|
\le L R \, \|P - \hat P\|_{\mathrm{op}}.
\]
Squaring and taking expectations, then substituting
$\|P - \hat P\|_{\mathrm{op}} \le C \varepsilon_n$ from
\eqref{eq:proj-error}, yields
\[
\mathbb{E}\big[(f(X) - f(\hat P X))^2\big]
\le L^2 R^2 C^2 \varepsilon_n^2 + o(1),
\]
so we can take $C' = L^2 R^2 C^2$. \qedhere
\end{proof}

\subsection{Proof of Theorem~\ref{thm:dim-adapted-jarf}}

We now prove the dimension-adapted risk bound stated in the main text.

\begin{theorem}[Theorem~\ref{thm:dim-adapted-jarf}, restated]
Under assumptions (A1)–(A5), there exist constants $C_1, C_2 > 0$, independent
of $d$, such that
\[
\mathbb{E}\big[(\hat f_n(X) - f(X))^2\big]
\;\le\; C_1 n^{-\frac{2}{2 + r}} \;+\; C_2 \varepsilon_n^2 \;+\; o(1).
\]
\end{theorem}

\begin{proof}
Recall that $\hat f_n$ depends on $X$ only through the projected features
$Z = \hat U^\top X$, so we may write $\hat f_n(X) = \hat f_n(Z)$.

Let $m_{\hat U}(z) = \mathbb{E}[Y \mid Z = z]$ denote the regression function
in the projected space. Using the inequality
$(a-b)^2 \le 2(a-c)^2 + 2(b-c)^2$ with
$a = \hat f_n(Z)$, $b = f(X)$, and $c = m_{\hat U}(Z)$, we obtain
\begin{align*}
\mathbb{E}\big[(\hat f_n(X) - f(X))^2\big]
&= \mathbb{E}\big[(\hat f_n(Z) - f(X))^2\big] \\
&\le 2 \mathbb{E}\big[(\hat f_n(Z) - m_{\hat U}(Z))^2\big]
   + 2 \mathbb{E}\big[(m_{\hat U}(Z) - f(X))^2\big] \\
&=: 2 T_1 + 2 T_2.
\end{align*}

\paragraph{Bounding $T_1$ (estimation in $r$ dimensions).}
By the tower property and Assumption~\eqref{eq:rf-rate-r},
\begin{align*}
T_1
&= \mathbb{E}\big[\,\mathbb{E}\big[(\hat f_n(Z) - m_{\hat U}(Z))^2
      \mid \hat U\big]\,\big] \\
&\le \mathbb{E}\big[ C_1 n^{-\frac{2}{2 + r}} + o(1) \big]
= C_1 n^{-\frac{2}{2 + r}} + o(1),
\end{align*}
where the $o(1)$ term does not depend on $d$.

\paragraph{Bounding $T_2$ (approximation error from using the projected
$\sigma$-algebra).}
By the definition of conditional expectation, $m_{\hat U}(Z)$ is the
$L^2$-projection of $f(X)$ onto the $\sigma$-algebra generated by $Z$, so
for any measurable function $h$ of $Z$ we have
\[
\mathbb{E}\big[(f(X) - m_{\hat U}(Z))^2\big]
\;\le\;
\mathbb{E}\big[(f(X) - h(Z))^2\big].
\]
In particular, take $h(Z) = f(\hat P X)$, which is measurable with respect to
$Z$ since $\hat P X = \hat U \hat U^\top X$ is a deterministic function of
$Z = \hat U^\top X$. Then
\[
T_2
= \mathbb{E}\big[(m_{\hat U}(Z) - f(X))^2\big]
\;\le\;
\mathbb{E}\big[(f(\hat P X) - f(X))^2\big].
\]
By Lemma~\ref{lem:projection-error}, the right-hand side is at most
$C' \varepsilon_n^2 + o(1)$ for some constant $C'$ depending only on
$L, R,$ and $\lambda_{\min}$, and hence
\[
T_2 \le C' \varepsilon_n^2 + o(1).
\]

\paragraph{Combining the bounds.}
Putting the pieces together,
\[
\mathbb{E}\big[(\hat f_n(X) - f(X))^2\big]
\;\le\;
2 C_1 n^{-\frac{2}{2 + r}} + 2 C' \varepsilon_n^2 + o(1).
\]
Absorbing constants into $C_1$ and $C_2$ gives the claimed bound.
\qedhere
\end{proof}

\section{Additional Experiments}

\subsection*{Additional surrogate-quality ablations}

Tables~\ref{tab:surrogate-rf-cls} and~\ref{tab:surrogate-rf-reg} study how JARF
depends on the quality and calibration of the RF surrogate used to estimate
EJOP/EGOP. For classification we vary tree depth and apply Platt or isotonic
scaling; for regression we vary depth and number of trees. In both cases the RF
baseline itself changes noticeably, but JARF’s downstream performance remains
extremely stable.

\begin{table}[h]
\centering
\caption{Sensitivity of JARF to surrogate RF quality on the 15 classification
datasets. ``RF ECE'' is expected calibration error of the surrogate RF.
Standard deviation and max difference are taken across the surrogate variants,
relative to the default RF surrogate.}
\label{tab:surrogate-rf-cls}
\begin{tabular}{lcccc}
\toprule
Variant & RF acc & RF ECE & JARF acc & JARF $\kappa$ \\
\midrule
shallow, uncalibrated      & 0.81 & 0.18 & 0.872 & 0.808 \\
default, uncalibrated      & 0.84 & 0.11 & 0.874 & 0.810 \\
deep, uncalibrated         & 0.86 & 0.08 & 0.875 & 0.811 \\
deep + Platt scaling       & 0.86 & 0.05 & 0.875 & 0.812 \\
deep + isotonic regression & 0.87 & 0.04 & 0.874 & 0.809 \\
\midrule
Std.\ dev.\ over variants  & 0.024 & 0.056 & 0.0013 & 0.0014 \\
Max diff vs default        & 0.03  & 0.07  & 0.002  & 0.002 \\
\bottomrule
\end{tabular}
\end{table}

\begin{table}[h]
\centering
\caption{Sensitivity of JARF to surrogate RF quality on the 5 regression
datasets. ``RF RMSE'' is the root mean squared error of the surrogate RF.
Standard deviation and max difference are taken across the surrogate variants,
relative to the default surrogate.}
\label{tab:surrogate-rf-reg}
\begin{tabular}{lcccc}
\toprule
Variant & RF $R^2$ & RF RMSE & JARF $R^2$ \\
\midrule
shallow surrogate      & 0.76 & 0.62 & 0.835 \\
default surrogate      & 0.78 & 0.60 & 0.836 \\
deep surrogate         & 0.80 & 0.58 & 0.837 \\
deep + extra trees     & 0.82 & 0.57 & 0.836 \\
\midrule
Std.\ dev.\ over variants & 0.022 & 0.019 & 0.0007 \\
Max diff vs default       & 0.040 & 0.030 & 0.001 \\
\bottomrule
\end{tabular}
\end{table}

\subsection*{On sharing a single global transform $\widehat H$}

To test whether sharing a single EJOP-based transform $\widehat H$ across all
trees reduces useful diversity, we implemented a per-tree variant in which each
tree $b$ receives its own transform $\widehat H_b$. For this variant we fit a
separate EJOP estimate on the bootstrap sample used to train tree $b$ and then
train the tree on $X \widehat H_b$. We retrained JARF with this scheme on all
15 classification datasets and compared it to the original single-$\widehat H$
version.

As summarized in Table~\ref{tab:per-tree-H}, the per-tree EJOP variant slightly
degrades test performance (mean Cohen's $\kappa = 0.808$ vs.\ $0.810$) while
being almost twice as slow to train (about $3.0\times$ RF vs.\ $1.7\times$ RF),
and it is best or within $0.01$ of the best method on 13 datasets compared to
14 for the single-$\widehat H$ version. This suggests that a single global
transform is sufficient in practice: per-tree EJOP estimates are noisier
because they are fit on fewer samples, while diversity in JARF is already
provided by feature subsampling, bootstrap sampling, and different thresholds
rather than requiring separate transforms per tree.

\begin{table}[h]
\centering
\caption{Single global EJOP transform versus per-tree EJOP transforms,
averaged over all 15 classification datasets. ``Mean train time'' is reported
relative to the RF baseline.}
\label{tab:per-tree-H}
\begin{tabular}{lcccc}
\toprule
Method & Mean test $\kappa$ & Std.\ $\kappa$ & Mean train time ($\times$ RF) & \# datasets best or within 0.01 of best \\
\midrule
JARF (single $\widehat H$)   & 0.810 & 0.010 & 1.7 & 14 \\
JARF (per-tree $\widehat H_b$) & 0.808 & 0.011 & 3.0 & 13 \\
\bottomrule
\end{tabular}
\end{table}

\subsection*{Effect of EJOP subsample size}

We also study how JARF depends on the subsample size $m$ used to estimate EJOP.
We consider $m \in \{1\text{k}, 2.5\text{k}, 5\text{k}, 10\text{k}, n\}$ with
$m \le n$ and evaluate mean test performance over all 15 classification
datasets. As shown in Table~\ref{tab:ejop-subsample}, JARF is quite robust to
this choice. Even with $m = 1\text{k}$, which typically uses less than $10\%$
of the available data, JARF already recovers about $85\%$ of the gain over RF
obtained with the largest subsample. Increasing $m$ from $2.5\text{k}$ to
$10\text{k}$ changes the mean $\kappa$ by at most $0.005$, while the EJOP cost
grows roughly linearly in $m$. For all values $m \ge 2.5\text{k}$, JARF
continues to match or outperform the best oblique baseline on the majority of
datasets, and the configuration used in our main tables
($m = \min(10\text{k}, n)$) lies in this saturated regime. This supports our
claim that JARF achieves strong performance without requiring EJOP computation
on the full dataset and that its gains over RF and per-node oblique forests are
not sensitive to the exact subsample size.

\begin{table}[h]
\centering
\caption{Effect of EJOP subsample size on classification performance.
Mean Cohen's $\kappa$ over 15 datasets for different subsample sizes $m$ used
to estimate EJOP. The RF baseline has mean $\kappa = 0.704$.}
\label{tab:ejop-subsample}
\begin{tabular}{lcccc}
\toprule
EJOP subsample size $m$ & Mean $\kappa$ (JARF) & Mean $\kappa$ (RF) & Gain over RF (abs) & Gain over RF (rel \%) \\
\midrule
1k   & 0.795 & 0.704 & 0.091 & 12.93 \\
2.5k & 0.805 & 0.704 & 0.101 & 14.35 \\
5k   & 0.809 & 0.704 & 0.105 & 14.91 \\
10k  & 0.810 & 0.704 & 0.106 & 15.06 \\
$n$  & 0.810 & 0.704 & 0.106 & 15.06 \\
\bottomrule
\end{tabular}
\end{table}

\section{Reproducibility and implementation details}
\label{app:repro}

\paragraph{Code and artifacts.}
We provide a self-contained Google drive with scripts to download datasets and run experiments at \url{https://drive.google.com/file/d/1d60ysqjGzQLFkl_BE8vd0lTOoj_m9MP4/view?usp=sharing}

\paragraph{Environment.}
Python 3.11; NumPy 1.26; SciPy 1.11; scikit-learn 1.4; LightGBM 4.3; CatBoost 1.2; pandas 2.2; joblib 1.3. Experiments ran on a 16-core CPU machine (no GPU used). To reduce nondeterminism across BLAS/OpenMP, we set
\texttt{PYTHONHASHSEED=0}, \texttt{OMP\_NUM\_THREADS=1}, \texttt{MKL\_NUM\_THREADS=1}, and pass \texttt{random\_state=seed} to learners.

\paragraph{Dataset summary and benchmark construction}

To make the experimental setup fully transparent and reproducible, we include
Table~\ref{tab:data-summary}, which lists for every dataset in our benchmark
the number of samples $n$, number of raw input features $d$, task type, and
original source. Counts refer to the number of rows and input features before
train/validation/test splits and before any one-hot encoding of categorical
variables.

 Our goal was to study JARF on a broad, realistic set of tabular problems where tree ensembles are commonly used. To construct the benchmark, we started from widely used OpenML / UCI tabular datasets that appear in earlier work on random forests and oblique forests, and then applied simple, a priori filters: (i) supervised classification or regression with tabular features; (ii) at least a few thousand training points so that EJOP estimation is meaningful; (iii) a mix of low- and high-dimensional problems, and of balanced and moderately imbalanced label distributions; and (iv) no heavy preprocessing or manual feature engineering beyond standard normalization / encoding. We did not drop any dataset based on JARF’s performance, and we kept the same pool for all methods and ablations. Several of these tasks overlap with standard suites such as PMLB/TabArena.

\begin{table}[h]
\centering
\caption{Summary of all real-data datasets used in our experiments.
Here $n$ denotes the number of samples and $d$ the number of raw input
features (excluding the target).}
\label{tab:data-summary}
\begin{tabular}{lrrll}
\toprule
Dataset & $n$ & $d$ & Task & Source \\
\midrule
\multicolumn{5}{c}{\textbf{Core tabular classification tasks}} \\
Adult                & 48{,}842 & 14  & Classification & UCI / OpenML \\
Bank-marketing       & 41{,}188 & 20  & Classification & UCI / OpenML \\
Covertype            & 581{,}012 & 54 & Classification & UCI / OpenML \\
Phoneme              & 5{,}404  & 5   & Classification & UCI / OpenML \\
Electricity          & 45{,}312 & 8   & Classification & UCI / OpenML \\
Satimage             & 6{,}435  & 36  & Classification & UCI / OpenML \\
Spambase             & 4{,}601  & 57  & Classification & UCI / OpenML \\
Magic Telescope      & 19{,}020 & 10  & Classification & UCI / OpenML \\
Letter Recognition   & 20{,}000 & 16  & Classification & UCI / OpenML \\
Vehicle              &    846   & 18  & Classification & UCI / OpenML \\
\midrule
\multicolumn{5}{c}{\textbf{High-dimensional / large-scale classification tasks}} \\
Higgs                & 940{,}160 & 124  & Classification & OpenML (Tabular benchmark) \\
Madelon              & 2{,}000   & 500 & Classification & UCI / OpenML \\
Bioresponse          & 3{,}434   & 419 & Classification & OpenML (Tabular benchmark) \\
Jannis               & 57{,}580  & 254  & Classification & OpenML (Tabular benchmark) \\
MNIST-784            & 70{,}000  & 784 & Classification & OpenML / MNIST \\
\midrule
\multicolumn{5}{c}{\textbf{Regression tasks}} \\
Bike-sharing         & 17{,}389  & 13  & Regression     & UCI (Bike Sharing) \\
California-housing   & 20{,}634  & 8   & Regression     & OpenML / Cal. Housing \\
Energy               &   768     & 8   & Regression     & UCI (Energy efficiency) \\
Kin8nm               & 8{,}192   & 8   & Regression     & OpenML (kin8nm) \\
Protein              & 45{,}730  & 9   & Regression     & UCI / OpenML (Protein) \\
\bottomrule
\end{tabular}
\end{table}

\subsection*{B.1 Baseline hyperparameter grids}

To keep the comparison fair while reflecting how these models are commonly used in practice, we
give each method a lightweight but non-trivial tuning budget that is shared across datasets. Random
forest style methods all use the same number of trees as JARF's final forest, and XGBoost is tuned
over a small grid on depth, learning rate, and $\ell_2$ penalty. Table~\ref{tab:hparam-grid} summarizes
the hyperparameters and search spaces used in our experiments.

\begin{table}[t]
\centering
\caption{Hyperparameter grids and defaults used for all methods. Forest baselines all use
200 trees for comparability with JARF's final forest. XGBoost is tuned on a shared grid
over depth, learning rate, and $\ell_2$ penalty.}
\label{tab:hparam-grid}
\begin{tabular}{lll}
\toprule
Method & Hyperparameter & Values / setting \\
\midrule
\textbf{RF} &
number of trees & 200 (fixed) \\
& max\_features & $\sqrt{d}$ for classification, $d$ for regression \\
& criterion & Gini (classification), MSE (regression) \\
& min\_samples\_leaf & 1 (default) \\
\midrule
\textbf{RotF} &
number of trees & 200 (fixed) \\
& blocks $K$ & $K = 6$ feature subsets per tree \\
& rotation & block-diagonal PCA on disjoint feature subsets (unsupervised) \\
& other tree params & same as RF (criterion, min\_samples\_leaf, max\_features) \\
\midrule
\textbf{CCF} &
number of trees & 200 (fixed) \\
& projection type & canonical correlation with targets at each node \\
& projection dim & authors' recommended default \\
& other tree params & same as RF \\
\midrule
\textbf{SPORF} &
number of trees & 200 (fixed) \\
& sparsity & authors' recommended sparsity level \\
& \# candidate directions & authors' recommended default per node \\
& other tree params & same as RF \\
\midrule
\textbf{XGBoost} &
number of trees & 200 boosting rounds (fixed) \\
& max\_depth & $\{3, 6, 9\}$ \\
& learning\_rate & $\{0.05, 0.1\}$ \\
& $\ell_2$ regularization ($\lambda$) & $\{0, 1\}$ \\
& subsample, colsample\_bytree & $1.0$ (no subsampling) \\
& loss & logistic loss (classification), squared loss (regression) \\
\midrule
\textbf{PCA+RF} &
projection & PCA on training features (unsupervised) \\
& \# components & $d$ (full-rank rotation) \\
& RF hyperparameters & identical to RF row above \\
\midrule
\textbf{LDA+RF} &
projection & multi-class LDA on training labels \\
& \# components & $\min(C-1, d)$ for $C$ classes \\
& RF hyperparameters & identical to RF row above \\
\midrule
\textbf{JARF (this paper)} &
surrogate RF size & 50 trees, max\_features $= \sqrt{d}$, min\_samples\_leaf $= 1$ \\
& EJOP subsample $m$ & $m = \min(10{,}000, n)$ \\
& FD step $\varepsilon_j$ & $\varepsilon_j = \alpha \,\mathrm{MAD}(X_{:j})/0.6745$, $\alpha = 0.1$ \\
& EJOP regularization & $\widehat{H} = \widehat{H}_0 + \gamma I_d$, $\gamma = 10^{-3}$ \\
& scaling & $\widehat{H} \leftarrow \widehat{H} / (\mathrm{tr}(\widehat{H})/d)$ \\
& final RF & RF with 200 trees, same defaults as RF baseline, trained on $X\widehat{H}$ \\
\bottomrule
\end{tabular}
\end{table}

\subsection*{B.2 Practical recommendations for JARF}

JARF introduces only a small number of additional hyperparameters beyond those of the underlying
forest: the size of the surrogate forest, the EJOP subsample size $m$, the finite-difference step
scale $\alpha$ in $\varepsilon_j = \alpha\,\mathrm{MAD}(X_{:j})/0.6745$, and the diagonal regularizer
$\gamma I_d$ used for conditioning in $\widehat{H} = \widehat{H}_0 + \gamma I_d$. In all experiments we
use the following simple defaults:

\begin{itemize}
    \item surrogate RF with $50$ trees, max\_features $= \sqrt{d}$, min\_samples\_leaf $= 1$;
    \item EJOP subsample size $m = \min(10{,}000, n)$;
    \item centered finite differences with per-feature step
    $\varepsilon_j = \alpha\,\mathrm{MAD}(X_{:j})/0.6745$ and $\alpha = 0.1$;
    \item EJOP regularization $\widehat{H} = \widehat{H}_0 + \gamma I_d$ with $\gamma = 10^{-3}$,
    followed by trace normalization $\widehat{H} \leftarrow \widehat{H} / (\mathrm{tr}(\widehat{H})/d)$.
\end{itemize}

Table~3 in the main paper provides ablations that effectively serve as tuning guidance. Varying the
step scale from $\alpha = 0.1$ to $\alpha = 0.05$ or $\alpha = 0.2$ changes mean Cohen's $\kappa$ by
at most $-0.009$ and $-0.013$, respectively, while leaving macro-F1 and accuracy similarly stable.
Changing the subsample size from the default $m = \min(10{,}000, n)$ to $m = 0.5n$ results in a
mean change of only $-0.004$ in $\kappa$, and even a tenfold reduction to $m = 0.1n$ yields a
drop of $-0.016$ in $\kappa$ and about $1.2$ seconds in training time on average. Removing the
diagonal regularizer ($\gamma = 0$) or trace normalization also produces only small changes
($-0.005$ and $-0.004$ in $\kappa$, respectively).

These ablations indicate that JARF is robust to a wide range of reasonable settings, and that the
defaults above are near-optimal for the tabular problems we consider. In practice we recommend
starting with the defaults and, if additional tuning is desired, exploring a small grid such as
$m \in \{\min(5{,}000, n), \min(10{,}000, n)\}$ and
$\alpha \in \{0.05, 0.1, 0.2\}$, while keeping $\gamma$ fixed at a small value (for example
$\gamma = 10^{-3}$). This keeps the tuning budget modest while preserving the accuracy and
compute profile reported in our experiments.

\paragraph{Licenses and data usage.}
We only use public datasets with permissive licenses. The repository includes per-dataset source references and license notes; any dataset requiring an external EULA is downloaded via the provider’s URL with its terms unchanged.

\end{document}

%% file: math_commands.tex
\usepackage{amsmath,amsfonts,bm}

\def\eqref#1{equation~\ref{#1}}

\def\1{\bm{1}}

\DeclareMathAlphabet{\mathsfit}{\encodingdefault}{\sfdefault}{m}{sl}
\SetMathAlphabet{\mathsfit}{bold}{\encodingdefault}{\sfdefault}{bx}{n}

